\documentclass[shortpaper]{clv2025}

\jvol{00}\jnum{00}\jyear{2025}

\usepackage{xcolor}
\definecolor{darkblue}{rgb}{0, 0, 0.5}
\usepackage{hyperref}
\hypersetup{colorlinks=true,citecolor=darkblue, linkcolor=darkblue, urlcolor=darkblue}
\usepackage{amsmath}
\usepackage{graphicx}
\usepackage{paralist}
\usepackage{booktabs}
\usepackage{amsmath}
\usepackage{amssymb}
\usepackage{colortbl}
\usepackage{comment}
\newcommand{\cg}[1]{\textcolor{blue}{#1}}
\renewcommand{\cg}[1]{#1}

\title{Unveiling Affective Polarization Trends in Parliamentary Proceedings}
\runningtitle{Unveiling Polarization Trends in Parliamentary Proceedings}

\runningauthor{Goldin et al.}

\author{Gili Goldin\thanks{Corresponding author}$^{,1}$, Ella Rabinovich$^{2}$, Shuly Wintner$^{1}$}

\affilblock{
    \affil{Department of Computer Science, University of Haifa\\\quad \email{gili.sommer@gmail.com}}
    \affil{The Academic College of Tel Aviv-Yaffo}
}

\date{}

\begin{document}
\maketitle
\begin{abstract}
Recent years have seen an increase in polarized discourse worldwide, on various platforms. We propose a novel method for quantifying polarization, based on the emotional style of the discourse rather than on differences in ideological stands. 
Using measures of \emph{Valence}, \emph{Arousal} and \emph{Dominance}, we detect signals of emotional discourse and use them to operationalize the concept of \emph{affective polarization}. 
Applying this method to a recently released corpus of proceedings of the Knesset, the Israeli parliament (in Hebrew),
we find that the emotional style of members of government differs from that of opposition members; and that the level of affective polarization, as reflected by this style, is significantly increasing with time.

\end{abstract}

\section{Introduction}
\label{sec:introduction}
In recent years, polarization trends have become increasingly evident worldwide, particularly on social media  \citep{ doi:10.1073/pnas.1804840115,demszky-etal-2019-analyzing,milbauer-etal-2021-aligning,doi:10.1080/23808985.2021.1976070} but also in parliamentary discourse  \citep{RePEc:bin:bpeajo:v:43:y:2012:i:2012-02:p:1-81, Sakamoto2017CrossnationalMO, https://doi.org/10.1111/jcms.13448}. 
While much of the attention has focused on ideological divides, some studies have also emphasized the emotional dynamics of polarization, termed \emph{Affective Polarization}. 
Ideological polarization focuses more on divisions and opposing viewpoints, whereas affective polarization focuses on the emotional characteristics of the discourse.
The term was first introduced by \citet{10.1093/poq/nfs038} as an alternative, more significant indicator definition of polarization (than ideological polarization). 
It refers to the phenomenon whereby individuals not only disagree with members of opposing political or ideological groups, but also develop distrust, dislike, and strong negative feelings toward them. Affective polarization is also defined as the tendency to view like-minded partisans positively and opponents negatively \citep{10.1093/poq/nfs038, lerman2024affectivepolarizationdynamicsinformation, https://doi.org/10.1111/pops.12955}. 
A recent \href{https://www.pewresearch.org/politics/2014/06/12/political-polarization-in-the-american-public/}{Pew Research Center report} found that ``Beyond the rise in ideological consistency, another major element in polarization has been the growing contempt that many Republicans and Democrats have for the opposing party. To be sure, disliking the other party is nothing new in politics. But today, these sentiments are broader and deeper than in the recent past''; 
and \citet{VanBavel2024} demonstrated that affective polarization significantly contributed to negative public health outcomes. These findings underscore the critical importance of studying and understanding the emotional aspects of polarization.

In light of the growing polarization witnessed worldwide, the main research question we address in this work is thus: has political discourse in Israel become more polarized over the years\cg{, and can such trends be assessed computationally}?
We focus on affective polarization, and we conjecture that we can assess it by computational analysis of the style of political discourse. 
We propose a computational operationalization of affective polarization, in terms of three emotional dimensions of language that can be computed from text: \emph{Valence}, \emph{Arousal} and \emph{Dominance} (VAD). 
We constructed a large lexicon of Hebrew words with VAD values for each word \cg{by manually reviewing and correcting the automatic translation of the English VAD lexicon of \citet{mohammad-2018-obtaining}}.
Following prior work on English \citep{DBLP:journals/corr/abs-2008-05713, rabinovich-etal-2020-pick}, we trained effective models for predicting these measures at the word- and sentence-level. 
We then applied the VAD models to determine the three  emotional dimensions of sentences in The \emph{Knesset corpus} \citep{Goldin2024TheKC}, which includes over 30 years of texts from the Israeli parliament proceedings.
Our main finding is that affective polarization, as reflected by these measures in the committee deliberations of the Israeli parliament, indeed increases with time.


To validate the quality of the VAD models we investigated the stylistic differences in the language used by government\footnote{We use the terms ``government'' and ``coalition'' interchangeably throughout this work, in line with the structure of the Israeli parliamentary system.}  vs.\ the opposition, using the same three measures. 
Based on much research in political science and communication, we anticipated that the language of the opposition would be more emotionally intense, since opposition parties often aim to challenge and criticize the policies and actions of the government. 
For the same reason, we also anticipated the language of coalition members to be more positive compared to the opposition members. In addition, we expected the government to use more assertive language, reflecting their authority and control. 
We show that these trends are clearly reflected in the VAD measures computed on our corpus, confirming that VAD measures can be used to assess the emotional color of speech.


%

The contributions of this work are manifold. First, we believe that we are the first to operationalize polarization in terms of the emotional style of the text, and specifically to use VAD measures to assess affective polarization. Second, we introduce and release an encoder LLM which was fine-tuned on the Knesset data in an unconventional way, which we trust would be a valuable resource for various analyses of parliamentary discourse in Hebrew. Third, we curated various language resources for Hebrew, including a large-scale VAD lexicon, a small dataset of sampled Knesset sentences which were manually annotated with VAD values, and the Knesset corpus annotated automatically for VAD by our models. Finally and most importantly, we use these resources to answer a pertinent question in political science. All the resources produced in this work, including data and code, are publicly-available and can be accessed here: \url{https://github.com/HaifaCLG/Polarization}.

\section{Related Work}

\textbf{Polarization} in political discourse has been the focus of extensive study in recent years. 
\citet{RePEc:bin:bpeajo:v:43:y:2012:i:2012-02:p:1-81} analyzed the evolution of political language in the U.S.\ Congress, finding that political language has become more polarized over time, with specific phrases becoming strongly associated with particular political parties. 
\citet{demszky-etal-2019-analyzing} built a comprehensive framework for studying linguistic aspects of polarization in social media, by studying tweets about mass shooting events. They defined polarization as partisan linguistic differences and established that there was substantial polarization between Democrats and Republicans in these tweets.
\citet{milbauer-etal-2021-aligning} used multilingual word embeddings and the semantic divergence they induce to characterize polarization between online communities on \emph{Reddit}. This method highlighted the complexity of ideological polarization in online discourse and revealed significant polarization trends within online communities. 
\citet{sinno-etal-2022-political} presented a novel approach to understanding political ideology and polarization in news media, using a multi-dimensional ideological spectrum across Economic, Social, and Foreign dimensions. 
\citet{frimer2022incivility} found a significant increase in uncivil and toxic language among members of the U.S. congress on Twitter between 2009 and 2019. The primary measure of incivility was PerspectiveAPI's ``toxicity'' score, which they assigned to 1.3 million tweets by members of the U.S. congress. Uncivil tweets tended to receive more approval and attention encouraging politicians to engage in greater incivility. This could be another indicator of rising polarization trends.  

In all these works, \emph{polarization} was mainly characterized by significant differences in political views between groups. We study political polarization from a different (and complementary) angle: we focus on how polarization manifests emotionally within the texts, and assess it using emotion analysis tools. 
A similar approach motivated
\citet{külz2022unitedstatespoliticianstone}, who investigated the tone of political debate in the United States based on a corpus which contains quotes attributed to U.S.\  politicians from online news sources. They used Linguistic Inquiry and Word Count (LIWC, \citet{pennebaker2001linguistic,Tausczik2010ThePM}) to assess the emotional content of the language and to quantify negative language. They found a significant and abrupt increase in the use of negative language coinciding with the start of Donald Trump's primary campaign and remaining elevated through the rest of the study period (2020). This increase in negativity was reflected across the political spectrum, confirming a significant increase in the negative tone of U.S.\ political discourse. We also explore emotional language in the context of political polarization, but we use a more sophisticated approach: We train regression models on sentence embeddings to predict and capture emotions, offering a context-aware prediction of emotions at the sentence level, rather than relying on the much simpler method of word counting with the LIWC dictionary.

\textbf{Affective polarization} was defined by \citet{10.1093/poq/nfs038}, who examined political polarization in the American public. They argued that affective polarization, or the degree to which partisans dislike each other, is a more appropriate test of polarization than ideological identity. Using data from various surveys, they demonstrated that both Democrats and Republicans increasingly dislike and harbor negative feelings toward the opposing party over time. 
Similarly,
\citet{https://doi.org/10.1111/lsq.12374} 
defined \emph{polarizing rhetoric} not merely by addressing divisive issues, but also by using language specifically designed to create division between the speaker---via identification with an in-group---and an out-group. They linked polarizing rhetoric to affective polarization, which they defined as increased negative feelings or antipathy towards opposing partisans. Polarizing rhetoric is seen as ``the language of affective polarization''. 


More recently,
\citet{https://doi.org/10.1111/pops.12955} explored how emotions relate to affective polarization in a qualitative research, using interviews with German radical-right voters to examine the way partisans report their emotions. 
%
\citet{lerman2024affectivepolarizationdynamicsinformation} demonstrated how affective polarization manifests emotionally. Using large-scale social media data, they analyzed replies of users with opposing ideologies. They measured the emotional tone of interactions by counting specific emotion-laden words, such as those expressing anger, joy, or disgust, and assessing toxicity levels. Their findings revealed that positive emotions, like joy, are more prevalent in in-group interactions, while out-group interactions are marked by anger, disgust, and toxicity. This evidence motivates us to explore affective polarization in parliamentary discourse, but whereas they relied on simpler word-count-based measures of emotion, we use contemporary tools and advanced models to predict nuanced sentence-level emotional dimensions.

\textbf{\emph{Valence}, \emph{Arousal}, and \emph{Dominance} (\emph{VAD})} are three dimensions, first introduced by \citet{mehrabian1974approach}, that are used to quantify emotions encoded in words. \emph{Valence} (or \emph{pleasure}) measures the positivity or negativity of an emotion. \emph{Arousal} measures the intensity of the emotion, ranging from calm to excited, and \emph{Dominance} indicates the degree of control, from submissive to dominant, encoded in a word. 
VAD is a common approach to measure emotion in language \citep{preotiuc-pietro-etal-2016-modelling, DBLP:journals/corr/abs-2008-05713, park-etal-2021-dimensional, wang-etal-2023-t2iat}.
Despite the word-level annotations, the measures have been successfully applied to longer text, e.g., sentences \citep{DBLP:journals/corr/abs-2008-05713, rabinovich-etal-2020-pick}. 
The VAD model is essential in sentiment analysis, providing a nuanced understanding of emotional expressions beyond simple positive or negative labels \citep{Bradley1994MeasuringET, mohammad-2018-obtaining}. Examples of VAD-scored sentences from the EmoBank dataset \citep{buechel-hahn-2017-emobank} are provided in Table~\ref{tbl:emo_bank_examples}. 

\begin{table*}[hbtp]
\resizebox{\textwidth}{!}{
\centering
\begin{tabular}{p{11cm}ccc}
\toprule
\textbf{Sentence} & \textbf{V} & \textbf{A} & \textbf{D} \\ 
\midrule

My girlfriend has disappeared, I don't even know where to start looking, and I need help! & 0.11 & 0.85 & 0.15 \\ 
\midrule
For a perfect moment, Emil and Tasha and I were one entity, laughing until our lungs hurt. & 0.94 & 0.80 & 0.56 \\ 
\midrule
We're at an exciting juncture and it should not take long until we start seeing solid results from our efforts.
& 0.76 & 0.73 & 0.83 \\ 
\midrule

But in fact, once news of the handover vanished from the front pages, the people of Hong Kong returned to their usual topics of conversation: the economy and the price of housing. & 0.5 & 0.12 & 0.63 \\ 
\bottomrule
\end{tabular}
}
\caption{Sentences from the EmoBank dataset \citep{buechel-hahn-2017-emobank} with their VAD scores. 
} 
\label{tbl:emo_bank_examples}
\end{table*}

VAD measures have been sporadically used to assess sentiments related to polarization. 
For example, \citet{vorakitphan-etal-2020-regrexit} used VAD to assess the sentiment toward various aspects of the \emph{Brexit} question which polarized the British society: the sentiment toward Brexit, but also towards immigration, a referendum, and politician names. 
\citet{lee-etal-2022-neus} used VAD to measure media bias in a corpus of political reports in the U.S.; \citet{upadhyaya-etal-2023-toxicity} used them to detect stance in Twitter posts, and \citet{shiwakoti-etal-2024-analyzing} to explore stance towards climate change discourse on Twitter. In all these works, the VAD values assigned to sentences are simply summed up (or averaged) over the words in the sentence; no attempt is made to operationalize the concept of polarization.

In light of the success of VAD metrics in capturing emotions in texts, we use them here to explore affective polarization in the Knesset proceedings. To the best of our knowledge we are the first to use VAD for affectional analysis of political discourse, and the first to use VAD for sentiment analysis in Hebrew.

\textbf{Parliamentary proceedings} attract the attention of researchers in language technology and in the social sciences, and corpora reflecting the deliberations of parliaments around the world have been compiled for dozens of languages and countries \citep[see discussions in][]{FIŠER18.14,Parla-CLARIN}. 
A large-scale European project, \emph{ParlaMint}, currently distributes some~30 corpora of parliamentary debates of various European countries; these corpora are meticulously encoded according to a uniform schema, undergo strict tests that guarantee their compatibility with the encoding standards, contain rich metadata about speakers, and are linguistically annotated with morpho-syntactic information and named entities and are also machine translated to English \citep{10.1007/s10579-021-09574-0, Erjavec2024}.  
Our work focuses on the \emph{Knesset Corpus} \citep{Goldin2024TheKC}, which is distributed on the ParlaMint platform and adheres to its standards; our results, therefore, can be straight-forwardly replicated not only for Hebrew but for a plethora of other languages.

\textbf{Israel} serves as a compelling case study of political polarization: On the one hand, studies have highlighted its deep ideological divisions and growing polarization \citep{amitai2023political, amsalem2024causal}, but while polarization in Israel is significant, it is not an extreme outlier compared to trends observed in other countries \citep{gidron2020american}. 
As far is we know, ours is the first work that uses language technology to identify polarization in Hebrew.

\section{Experimental Setup and Methodology}
\label{sec:methodology}
\begin{figure*}[hbt]
\centering
\includegraphics[width=0.90\textwidth]{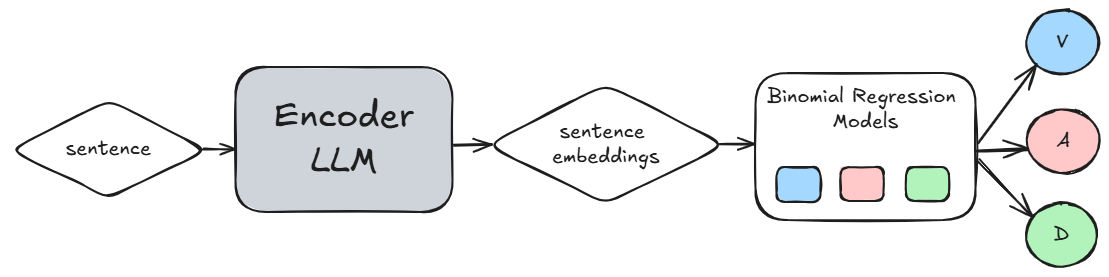}
\caption{VAD prediction combined model flow. The Encoder LLM receives a sentence as input and generates its sentence embedding. 
This embedding is then used as input for the regression models, which predict the Valence (V), Arousal (A) and Dominance (D) scores for the sentence. 
}
\label{fig:flow}   
\end{figure*}

We use the Knesset Corpus \citep{Goldin2024TheKC} 
as our primary source of data (Section~\ref{sec:data}). Our main vehicle for operationalizing affective polarization is the VAD measures of emotion: We characterize polarization using VAD and compute various metrics based on these VAD values. 
In order to assess emotional trends in text with VAD, and use them to compute polarization, we need VAD lexicons of Hebrew, as well as datasets annotated with these values; 
Section~\ref{sec:data_preprocessing} describes how we created these resources. Next, we discuss in Section~\ref{sec:models} regression models for predicting the VAD values of sentences in the corpus. Since these regression models obtain sentence embeddings as input, we utilized a Hebrew encoder LLM to associate each sentence in the corpus with its sentence embeddings (Section~\ref{sec:multi_knesset_model}). Then, we trained the regression models that obtain these sentence embeddings as input and predicted VAD values for each sentence (Section~\ref{sec:vad_binom_models}). Finally, we applied these models (Section~\ref{sec:eval_combined_models}) and assigned VAD values to every sentence in our corpus (see Figure~\ref{fig:flow}).
We conducted a qualitative assessment of the resulting VAD assignments, which we describe in Section~\ref{sec:qualitative-eval}.

\subsection{Data}
\label{sec:data}
Our main dataset is the Knesset corpus \citep{Goldin2024TheKC}, a large corpus of Hebrew parliamentary proceedings containing over~30 million sentences. The corpus includes both plenary sessions (from the years 1992-2024) and committee deliberations (1998-2024) held in the Israeli parliament. The sentences in the corpus are annotated with detailed meta-information reflecting political properties of the speakers. Each sentence is associated with its speaker and is linked to the relevant Knesset member, if applicable, together with their political meta-information such as faction affiliations and their government vs.\ opposition membership at that time. The sentences are also labeled with the session type (plenary or committee) and the committee name, where applicable. 

Since the plenary protocols mostly consist of prepared and edited speeches,
they are less likely to reflect extremely emotional and polarized discourse. In contrast, committee protocols consist mainly of spontaneous spoken language and are more likely to reveal polarization, if it exists. 
Furthermore, the topic of a discussion can largely affect the VAD values. For 
example, security issues are likely to yield discussion with high values of Dominance, whereas welfare discussions may display low Valence values.
There are also certain topics that are controversial and therefore more likely to cause disagreement or a heated argument. As we are interested in the stylistic differences between government and opposition, and in trends that may occur over time, we do not want the topics to interfere with our measurements.

We therefore focus in this study on the committee protocols and leave the plenary session protocols for future research.
To avoid confounds related to the topic, we calculate the VAD measures for each committee independently. 
Since each Knesset committee is dedicated to a specific issue, we grouped together all the sentences of a given committee as related to the same topic. Some committees changed their names throughout the years; we still grouped them together, based on our understanding of the topics discussed in their deliberations. We experimented only with committees whose proceedings consisted of more than~400K sentences, as many of the ``smaller'' committees are ad-hoc and do not span over all the years of the corpus. 
Table~\ref{tbl:committee_list} provides statistics on these committees.

\begin{table*}[hbtp]
\centering
\begin{tabular}{lrrr}
\toprule
\textbf{Committee Name} & \textbf{Protocols} & \textbf{Sentences} & \textbf{Tokens}  \\ 
\midrule
Finance         &  5699 & 3,587,061    & 36,504,569                     \\ 
Constitution, Law and Justice & 4461           & 3,391,760            & 38,992,684             \\ 
Economic Affairs          & 5103  & 3,372,319    & 36,953,390                      \\ 
Internal Affairs and Environment   & 3918         & 2,647,758        & 28,875,584                 \\ 
Labor, Welfare and Health & 4832   &  2,527,536       & 28,800,611                         \\ 
Education, Culture and Sports & 3851 & 2,341,827 & 28,486,694\\ 
State Control & 1802 & 1,290,939 & 16,126,994\\
Status of Women and Gender Equality &1511 & 941,561 & 13,059,157\\ 
Immigration, Absorption and Diaspora & 1753  & 936,962 & 12,384,494\\ 
House & 2182 & 837,049 & 8,976,718\\ 
Rights of the Child & 1071 & 768,400 & 9,903,766\\ 
Science and Technology & 836 & 601,223 & 8,242,832\\ 
Public Petitions & 837 & 532,772 &  6,449,128\\ \midrule
Total & 37,856 &23,777,167 & 273,756,621
\\
\end{tabular}
\caption{Dataset statistics.}
\label{tbl:committee_list}
\end{table*}

\subsection{Creating Hebrew VAD Datasets}
\label{sec:data_preprocessing}
To the best of our knowledge, no manually curated VAD lexicons existed for Hebrew; we therefore adapted the existing VAD lexicon of \citet{mohammad-2018-obtaining}, containing 19,972 English entries (Section~\ref{sec:vad_lexicon}). Additionally, we translated and used the EmoBank dataset \citep{buechel-hahn-2017-emobank}, as described in Section~\ref{sec:emobank}. In order to fit the models to work well with our domain, we also annotated a subset of the Knesset corpus for VAD values (Section~\ref{sec:manual_vad_annot}), and used part of it as additional data for training, and the other part for evaluation of the models.

\subsubsection{A Hebrew VAD Lexicon}
\label{sec:vad_lexicon}
The VAD lexicon \citep{mohammad-2018-obtaining} was originally created in English and was automatically translated and published in other languages, including Hebrew. Recognizing the inherent limitations of automatic translations, we observed inaccuracies and errors in the Hebrew version. To improve the reliability of this resource, we manually reviewed and corrected the translations. 

The original translations were written with diacritics, in what is known as \emph{dotted Hebrew}, while the text of most Hebrew texts, including the Knesset proceedings, is written without diacritics \citep{semitic-introduction,Itai2008LanguageRF}. Consequently, we employed three highly qualified in-house annotators (two women, one man, native Hebrew speakers with degrees in Linguistics) who converted each translation to an \emph{undotted} (non-diacriticized) form, in alignment with Hebrew orthographic rules. They also enriched each word in the lexicon by adding its lemma, as well as  alternative translations and synonyms in Hebrew.%
\footnote{\cg{The instructions for the annotators were: ``In case the (automatic) translation is inaccurate, correct  it. When the English word is ambiguous or has multiple translations in Hebrew, add all translation equivalents. When the Hebrew translation is dotted, add also the undotted form of the word. If the original translation is an inflected form, add also its lemma.''
}}
Each Hebrew word, lemma and other enrichments were assigned the same VAD scores as the original English word in the lexicon. 
We evaluated the agreement among our three annotators by selecting~400 lexical items and computing the percentage of items in which at least one of the translations was identical for all annotators, and the percentage of items in which all annotators agreed on the lemma. The results were~71.17\% and~84.18\%, respectively.

\paragraph{Validating the Hebrew VAD lexicon values}
Admittedly, potential cultural differences across languages may hamper the accuracy of VAD values of the translated words; however, most affective norms have been shown to be stable across languages, despite some cultural differences \cite{mohammad-2018-obtaining}. 
%
For example, automatically translated VAD metrics were used to recognize emotions in Spanish texts \citep{10.3389/fpsyg.2022.849083}, evaluate emotions towards the COVID19 pandemic in Arabic \citep{arabic-vad}, analyze dehumanization in Slovene \citep{caporusso-etal-2024-computational}, etc.
Furthermore, \citet{cho-etal-2025-language} demonstrated that Valence and Arousal values automatically translated to Mandarin were effective, in spite of cross-cultural differences between the U.S\ and China.

To further validate the Hebrew VAD scores we evaluated a sample of the lexicon as follows. 
We randomly selected 100 Hebrew terms for each VAD dimension, ensuring coverage across the full value range by selecting an equal number of entries from each value bracket. We selected only unambiguous Hebrew terms, appearing only once in the lexicon as a translation of an English term. 
Three annotators then annotated this subset for VAD, using the ranking-based scheme suggested by \citet{mohammad-2018-obtaining} (see Section~\ref{sec:manual_vad_annot} for the details).
These rankings were subsequently converted to continuous VAD scores.

We assessed inter-annotator agreement using mean pairwise Pearson correlation, showing correlations of 0.87, 0.67, and 0.65 for Valence, Arousal, and Dominance, respectively. We then averaged the three values for each term and VAD dimension and calculated Pearson correlation between our average scores and the lexicon scores, automatically inferred from English values. The results, 
0.802, 0.701, and 0.673 for Valence, Arousal, and Dominance, respectively,
indicate \cg{} strong correlations for Valence and somewhat lower but still robust correlations for Arousal and Dominance. This experiment demonstrates that our lexicon scores are reliable for Hebrew as well.


\subsubsection{Adaptation of the EmoBank Dataset}\label{sec:emobank}
In order to enrich the regression models’ training data with VAD-annotated \emph{sentences}, beyond words, we used the EmoBank dataset \citep{buechel-hahn-2017-emobank}, a corpus of 10K English sentences manually annotated for VAD values. We automatically translated this dataset to Hebrew using the \href{https://huggingface.co/google/madlad400-3b-mt}{Google/madlad400-3b-mt}
model \citep{kudugunta2023madlad400multilingualdocumentlevellarge}. We selected for each VAD dimension only sentences whose values are above~0.7 and under~0.3, since classifiers typically learn clearer separations when boundary data points are available for training. 
We chose these thresholds because they provided a good balance between selecting the best examples to avoid confusing the model and still having enough samples for training. We also only used sentences longer than~10 tokens and shorter than~30 since those resembled a typical sentence length.
The final number of EmoBank sentences we used was 242, 165 and 163 for V, A and D, respectively.

\subsubsection{Manual Annotation of Sampled Knesset Sentences with VAD Values}
\label{sec:manual_vad_annot}
We collected 120 short texts (sentences and brief paragraphs) from the Knesset corpus to assess and improve our models for predicting VAD values on Knesset data. 
We selected instances that we thought exhibited emotional content in at least one aspect of the VAD dimensions. This was important since randomly selecting sentences would probably result in mostly neutral texts, given the predominance of emotionally neutral language in political discourse.

For annotating the emotional content of these texts, we followed the word-level VAD method of \citet{mohammad-2018-obtaining}, adapting it for short text-level, as it has been shown that ranking is easier for humans than assigning a continuous score \citep{louviere1991best, cohen2003maximum, Louviere_Flynn_Marley_2015}. Leveraging the tuple-generating script of \citet{mohammad-2018-obtaining}, we generated 240 tuples, each consisting of four short texts. Our three in-house annotators annotated each tuple: they were requested to mark the text with the highest and lowest values for each VAD dimension. We further applied the scoring script of \citet{mohammad-2018-obtaining} to the annotations, to convert comparative scores (highest, lowest), into continuous VAD scores in the [-1, 1] range, then averaged the scores of the three annotators and normalized them to the [0, 1] range. 
These scores serve as the gold standard for our model evaluation. Some examples of texts from the sampled Knesset dataset with their annotated mean V, A and D scores are shown in Table~\ref{tbl:knesset_vad_sents} (to facilitate reading, we translated these texts from Hebrew to English using GPT-4 \citep{openai2023gpt4}). 

We evaluated inter-annotator agreement by computing the {mean pair-wise Pearson correlation} between individual annotator scores, after applying the scoring script. Despite the rather subjective nature of this task, we obtained reasonably good inter-annotator agreement scores: 0.904, 0.662 and 0.675 for V, A and D, respectively.

\begin{table*}[htbp]
\centering
\resizebox{\textwidth}{!}{
\begin{tabular}{p{12.5cm}ccc}
\toprule
\textbf{Text} & \textbf{V} & \textbf{A} & \textbf{D} \\
\midrule
I believe we will achieve an impressive success thanks to the committee -- the bill will finally bring transparency, equality, and the elimination of discrimination. & 0.979 & 0.396 & 0.917 \\
\midrule
We are living today in a blessed era, an era where there are empowered and empowering women, an era where there is absolute freedom of expression, an era of flourishing democracy, but most importantly, an era where we have a home. & 1.000 & 0.771 & 0.833 \\
\midrule
Do you remember what he shouted? What are you doing? Have you lost your mind? Get a grip, we have a nation that is dying. This is a matter of life and death. People are dying, and don't you dare blame anyone. & 0.000 & 0.958 & 0.271 \\
\midrule
I must note that even in previous Knessets, there were collaborations between the coalition and the opposition, and the issue of women's status crossed borders and political ideologies, but in this Knesset, particularly because of the special political constellation within the coalition, we are breaking new ground. & 0.958 & 0.604 & 0.917 \\
\midrule
It turns out this didn't start now; the lack of governance is now glaring in its full ugliness. & 0.292 & 0.479 & 0.146 \\
\bottomrule
\end{tabular}
}
\caption{Example Knesset texts with their (averaged) manually-annotated VAD scores.}
\label{tbl:knesset_vad_sents}
\end{table*}

\subsubsection{Summary of Resources: Hebrew VAD lexicons and datasets}
Summing up this preparation work, we have created (and released) the following resources:
\begin{itemize}
    \item An English-Hebrew VAD lexicon with 19,972 terms in English that are mapped to their (potentially multiple) Hebrew  translations and Hebrew lemmas. The number of unique Hebrew lexical terms is~13,526.\footnote{Due to the ambiguity of the Hebrew writing system, many Hebrew word forms can be translated to multiple English words.}
    \item A subset of the English EmoBank dataset automatically translated to Hebrew with 242, 165 and 163 sentences annotated for~V, A and D, respectively.
    \item A sample of~120 short texts from the Knesset manually annotated for VAD. 
\end{itemize}

\subsection{Models}
\label{sec:models}
A linear regression model (or its binomial variant \citep{Seabold2010StatsmodelsEA}, restricted to the [0--1] range) is an intuitive choice for the task of predicting a continuous score along each one of the VAD dimensions. The model receives numerical text representations; we therefore represent words and sentences with their embeddings. We first describe the LLM that we use to extract these embeddings (Section~\ref{sec:multi_knesset_model}) and then discuss the binomial regression models, which are designed to predict the VAD scores based on the embeddings (Section~\ref{sec:vad_binom_models}). We evaluate the end-to-end, combined flow in Section~\ref{sec:eval_combined_models}. The VAD prediction procedure is illustrated in Figure~\ref{fig:flow}.

\begin{figure}[hbt]
\centering
\includegraphics[width=0.70\columnwidth]{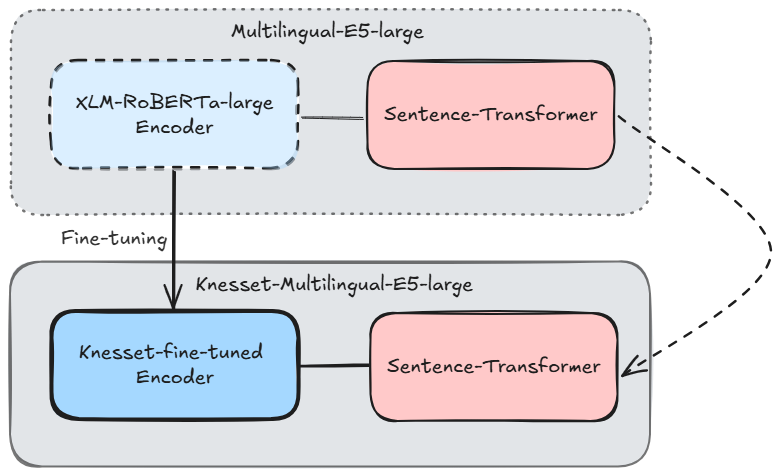}
\caption{The \texttt{Knesset-multi-e5-large} model, the chosen LLM for extracting sentence embeddings (see Figure~\ref{fig:flow}). It was created by fine-tuning the encoder part of the \texttt{multilingual-e5-large} model on the Knesset data, and then combining the tuned encoder with the \emph{original} sentence-transformer.}
\label{fig:model}   
\end{figure}

\subsubsection{The \texttt{Knesset-multi-e5-large} Language Model}
\label{sec:multi_knesset_model}
In order to obtain high quality sentence embeddings that can represent the meanings of the sentences in our data and  serve as reliable input for the VAD binomial models, 
an appropriate Hebrew encoder LLM is needed. We tested various off-the-shelf models (see Appendix~\ref{app:off-the-shelf-eval-res}), which we evaluated extrinsically on the final objective of predicting VAD scores, using the binomial models we describe in Section~\ref{sec:vad_binom_models}. 
Following this evaluation we decided to use the \texttt{multilingual-e5-large} model \citep{wang2024multilinguale5textembeddings} and fine-tune it on our Knesset data.

The \texttt{multilingual-e5-large} model is a sentence transformer model which consists of two main components: an \texttt{XLM-RoBERTa-large} encoder \citep{DBLP:journals/corr/abs-1911-02116} and a sentence-transformer component. The model was trained on the sentence similarity task \citep{cer-etal-2017-semeval}. In the absence of labeled Knesset data for this task, our initial attempt involved fine-tuning only the encoder component, using the Masked Language Modeling (MLM) task, on the Knesset data (32 million sentences), and using the representations from the last hidden layer as embeddings. This approach performed worse than the original \texttt{multilingual-e5-large} model. Recognizing that the sentence-transformer component contributed to the original model's success, we applied a non-trivial approach to fine-tune the model. We fine-tuned the encoder on the Knesset data using the MLM task, as before, and then used the output of this domain-adapted component as input to the original sentence-transformer component, as depicted in Figure~\ref{fig:model}. 
We thus manage to both fit the model to our data and keep the advantages of the pre-trained sentence transformer model. This approach moderately improved the results, compared to the original multilingual model, as shown in Section~\ref{sec:eval_combined_models}.
The full technical details of the training process are described in Appendix~\ref{app:finetune_training}.

\subsubsection{The VAD Binomial Regression Models}
\label{sec:vad_binom_models}
We trained a \href{https://www.statsmodels.org/stable/glm.html}{Generelized Linear Model (GLM) binomial model} \citep{Seabold2010StatsmodelsEA}\footnote{We used this Python \href{https://www.statsmodels.org/dev/glm.html}{GLM implementation}.} for each of the VAD dimensions. The Binomial family (with its default \texttt{logit()} link function) is a common choice in cases where the predicted value is a proportion, i.e., bounded by 0 and 1. For simplicity, we refer to this regression model as \emph{binomial model} hereafter. In our case, these binomial models receive sentence embeddings as input and predict the relevant dimension score of these sentences in the [0, 1] range, following the approach used in \citet{DBLP:journals/corr/abs-2008-05713} and \citet{rabinovich-etal-2020-pick} for English. We extracted the sentence embeddings of the labeled data that were described in Section~\ref{sec:data_preprocessing}, using the \texttt{Knesset-multi-e5-large} model described in Section~\ref{sec:multi_knesset_model}  to train the models.

\subsubsection{Evaluation of the Flow: Combined LLM and Regression Models}
\label{sec:eval_combined_models}
We compared the suggested approach to a similar flow, while using other off-the-shelf encoder models, including \texttt{DictaBert} \citep{shmidman2023dictabertstateoftheartbertsuite}, \texttt{alephBert} \citep{seker-etal-2022-alephbert}, \texttt{alephBertGimmel-base-512} \citep{gueta2023largepretrainedmodelsextralarge}, \texttt{Knesset-DictaBERT} \citep{goldin2024knessetdictaberthebrewlanguagemodel}, \texttt{multilingual-e5-large} \citep{wang2024multilinguale5textembeddings}, and others. We assessed their performance on the downstream task of predicting VAD scores using the binomial regression models described in  Section~\ref{sec:vad_binom_models}, using the Hebrew VAD lexicon we created. 
We used Pearson's correlation coefficient for evaluation, due to its effectiveness in measuring the linear correlation between variables. While all results were relatively close, the multilingual model yielded the best results for~A and~D, competitive results for~V, and the highest average performance overall. 
\texttt{alephBert} and \texttt{alephBertGimmel} performed worst and were therefore discarded. More details can be found in Appendix~\ref{app:off-the-shelf-eval-res}.

We further evaluated the performance of the selected models:  \texttt{DictaBert} \citep{shmidman2023dictabertstateoftheartbertsuite}, \texttt{Knesset-DictaBERT} \citep{goldin2024knessetdictaberthebrewlanguagemodel} and \texttt{multilingual-e5-large} \citep{wang2024multilinguale5textembeddings} by training them on all the available datasets described in Section~\ref{sec:data_preprocessing}: the Hebrew VAD lexicon, the translated EmoBank dataset, and a portion of the annotated Knesset sentences, while testing on the Knesset sentences not used in training. We evaluated the models using both a train-test split and 5-fold cross validation. For the train-test split, we used the VAD lexicon and the EmoBank dataset as training data, along with 70\% of the annotated Knesset sentences, leaving the remaining 30\% for testing. For the 5-fold cross validation, the full VAD lexicon and EmoBank dataset remained static in the training set, while the annotated Knesset sentences were split between the training and test sets for each fold. 

Table~\ref{tbl:pearson_vad_results} reports the evaluation results. The \texttt{multilingual-e5-large} model outperformed the other models in most cases, confirming its suitability as the best model for this task among the \emph{pre-trained} models. However, the results were still not satisfying, with Pearson's $r$ below 0.5 for A and D.\footnote{\cg{Correlations of $r > 0.5$ are considered ``large'', following conventions for effect size interpretation \citep{cohen1988statistical}, and we adopt this threshold in our analyses.}} Consequently, we fine-tuned this model on our data, producing the \texttt{Knesset-multi-e5-large} model. This process, described in Section~\ref{sec:multi_knesset_model}, produced the best model for our task, as shown in Table~\ref{tbl:pearson_vad_results}. 

\begin{table*}[htb]
\centering
\begin{tabular}{lcccccc}
\toprule
{} & \multicolumn{3}{c}{train-test split} & \multicolumn{3}{c}{5-fold} \\
\cmidrule(lr){2-4} \cmidrule(lr){5-7}
model name & V & A & D & V & A & D \\
\midrule
\texttt{dictaBERT} & 0.661 & 0.427 & 0.423 & 0.528 & 0.473 & 0.416 \\
\texttt{Knesset-dictaBERT} & 0.526 & 0.244 & 0.478 & 0.406 & 0.361 & 0.424 \\
\texttt{multilingual-e5-large} & 0.737 & 0.408 & 0.496 & 0.697 & 0.465 & 0.602 \\
\midrule
\texttt{\textbf{Knesset-multi-e5-large}} & \textbf{0.764} & \textbf{0.468} & \textbf{0.669} & \textbf{0.721} & \textbf{0.523} & \textbf{0.615} \\
\bottomrule
\end{tabular}
\caption{VAD prediction: Pearson's $r$ correlation  between the models' scores and the annotators' scores
, where the models were trained on all datasets and evaluated on a subset of Knesset sentences. Best results in each column are boldfaced.}
\label{tbl:pearson_vad_results}
\end{table*}

As an alternative approach to the one described above,
we also trained a transformer-based encoder model with a regression head (a single FFNN with one output unit, trained with an MSE objective) directly on the VAD-labeled data.
We used the encoder component of the same models we experimented with in Section~\ref{sec:eval_combined_models}, and attached to it a linear regression head layer. 
We then fine-tuned these models on the VAD data, using the same configuration as above for the train and test sets, along with a 5-fold cross-validation approach. The full technical details are presented in Appendix~\ref{app:vad_regression_head_training}. 

The results of this experiment are listed in Table~\ref{tbl:pearson_vad_results_regression_head}. While some models yielded better results for certain VAD values, neither model produced significantly better results overall compared to our original approach (see Table~\ref{tbl:pearson_vad_results}). We therefore use the binomial regression models described in Section~\ref{sec:vad_binom_models} in this work.

\begin{table*}[htb]
\centering
\begin{tabular}{lcccccc}
\toprule
{} & \multicolumn{3}{c}{train-test split} & \multicolumn{3}{c}{5-fold} \\
\cmidrule(lr){2-4} \cmidrule(lr){5-7}
model name & V & A & D & V & A & D \\
\midrule
\texttt{dictaBERT} & 0.440 & \textbf{0.633} & -0.046
& 0.269 & 0.327 & 0.058 \\
\texttt{Knesset-dictaBERT} & 0.150
& 0.212
& -0.268
& 0.081 & 0.108 & -0.123 \\
\texttt{multilingual-e5-large} & \textbf{0.749} & 0.333 & \textbf{0.536} & \textbf{0.690} & 0.275 & 0.496 \\
\midrule
\texttt{\textbf{Knesset-multi-e5-large}} & 0.521 & 0.475 & 0.446 & 0.630 & \textbf{0.361} & \textbf{0.498} \\
\bottomrule
\end{tabular}
\caption{VAD prediction: Pearson's $r$ correlation  between the regression-LLMs' scores and the annotators' scores. 
Best results in each column are boldfaced. 
Compare these results to Table~\ref{tbl:pearson_vad_results}.}
\label{tbl:pearson_vad_results_regression_head}
\end{table*}

Finally, we applied both the \texttt{Knesset-multi-e5-large model} and the binomial regression models to annotate all the sentences in our Knesset dataset with VAD values. 
The average values of~V,~A and~D were~0.51, 0.38 and~0.53, respectively. The distribution of `extreme' values%
\footnote{The range of VAD values is~0 to~1. We define `high' values as greater than~0.7, and `low' as lower than~0.3. These thresholds were determined arbitrarily, to strike a balance between focus on relatively extreme utterances, on one hand, and a substantial number of utterances that would guarantee the statistical significance of the results on the other. We also experimented with values of~0.9 for `high' and~0.1 for `low', and the results we report in Section~\ref{sec:validation-VAD} were very similar.}
was uneven, as we show in Table~\ref{tbl:corpus_vad_stats}.

\begin{table*}[hbtp]
\centering
\begin{tabular}{lrrr}
\toprule
VAD Statistic & \multicolumn{1}{c}{V} & \multicolumn{1}{c}{A} & \multicolumn{1}{c}{D}\\
\midrule
High & 12.31\%  &  0.59\% & 12.15\%  \\ 
Low  & 8.26\%   & 21.22\%  & 4.07\% \\ 
\bottomrule
\end{tabular}
\caption{Percentage of sentences with high and low VAD values in our dataset.}
\label{tbl:corpus_vad_stats}
\end{table*}



\subsection{Qualitative Assessment of the VAD Scores of Corpus Sentences}
\label{sec:qualitative-eval}
To further validate the assignment of VAD values to sentences in our corpus, as described in Section~\ref{sec:models}, we extracted the 20 highest- and lowest-valued sentences for each VAD metric in the protocols of each committee. Inspection of these sentences revealed that for Valence, sentences with high scores were evidently positive in tone, without exceptions, whereas those with low scores were always clearly negative.
The results of the Arousal assignments also aligned well with our expectations, though they were somewhat less pronounced than the case of Valence.
As for Dominance, the extracted sentences with the most extreme values appeared reasonable, but in some cases it was hard for us to understand why a particular sentence was assigned an extreme value. We note, however, that Dominance is not used in our operationalization of polarization (Section~\ref{sec:vad_over_time}).
These findings provide additional validation for our method of predicting VAD values for Knesset sentences.

For illustration, we present examples of Valence highest- and lowest-valued sentences from the protocols of the Economy Committee in Table~\ref{tbl:high_and_low_v_economy} (we translated the sentences to English for readability); Tables~\ref{tbl:high_and_low_a_economy} and~\ref{tbl:high_and_low_d_economy} depict examples of extreme Arousal and Dominance sentences, respectively. 
The full set of extremely-valued sentences for each VAD metric and committee is distributed with \href{https://github.com/HaifaCLG/Polarization/}{our supplementary materials}.

\begin{table*}[hbtp]
\centering
\begin{tabular}{p{0.8\textwidth}r}
\toprule
\textbf{Sentence} & \multicolumn{1}{c}{\textbf{V}} \\ 
\midrule
I am getting thousands of very-very positive responses.            & 0.950                      \\ 
Here we hear a positive answer, optimistic.    & 0.950            \\ 
To my delight, I have reached understandings with the ministry of economy and the ministry of communication, regarding a dramatic improvement in this matter.      & 0.949                     \\ 
We wish you, of course, congratulations.           & 0.947                      \\ 
I received it, it is very pretty.  & 0.947             \\
\midrule
The situation was very bad.            & 0.046                  \\ 
Worse than that.    & 0.051       \\ 
He didn't stand and people got hurt.      & 0.054                     \\ 
The situation was  catastrophic.          & 0.056                      \\ 
It was simply horrible.  & 0.056             \\
\bottomrule
\end{tabular}
\caption{Sentences with highest and lowest Valence values in the Economy Committee.}
\label{tbl:high_and_low_v_economy}
\end{table*}

\begin{table*}[hbtp]
\centering
\begin{tabular}{p{0.8\textwidth}r}
\toprule
\textbf{Sentence} & \multicolumn{1}{c}{\textbf{A}} \\ 
\midrule
If I charge and get killed, are we all a bunch of suicidal people?    & 0.874                 \\ 
I think they should arrive after us, as fast as possible of course, but first let us save lives, let us protect the living, that is the most important thing.           & 0.874                      \\ 
Whoever is a criminal should be punished.  & 0.947             \\
We have been warning, we are shouting out. & 0.872\\
\midrule
In Israel this number is currently 0.039.    & 0.053       \\ 
In the last year the prices of vegetables were reduced by 4.5\%, on average.      & 0.056                     \\ 
The average of investment was about 46 Shekels and 90 percent of the forms are 64 Shekels or less.        & 0.058                      \\ 
Regarding the glass, there is no use of green glass in Israel. & 0.062           \\
\bottomrule
\end{tabular}
\caption{Sentences with highest and lowest Arousal values in the Economy Committee.}
\label{tbl:high_and_low_a_economy}
\end{table*}

\begin{table*}[hbtp]
\centering
\begin{tabular}{p{0.8\textwidth}r}
\toprule
\textbf{Sentence} & \multicolumn{1}{c}{\textbf{D}} \\ 
\midrule
General Manager of the Ministry of Religious Affairs and acting General Manager of the Main Rabbinate of Israel.            & 0.943                     \\ 
Maximum implementation of the results of investments, research, and development.   & 0.937      \\ 
Sport industries that the committee believes the public has interest in?         & 0.926                     \\ 
It was published today, and there is room to increase the competition.  & 0.926             \\
\midrule
They don't pay attention for five minutes to 20 children sleeping on the street.          & 0.097                 \\ 
The weak were hurt here.   & 0.099       \\ 
Some of the disabled are not getting.      & 0.100                     \\ 
It only shows how bad our situation is.        & 0.101                     \\ 
I don't know where I am. & 0.104         \\
\bottomrule
\end{tabular}
\caption{Sentences with highest and lowest Dominance values in the Economy Committee.}
\label{tbl:high_and_low_d_economy}
\end{table*}

\section{Validation of the VAD measures}
\label{sec:validation-VAD}

To further validate the adequacy of the VAD measures as a vehicle for assessing emotional trends in the style of political proceedings, we describe in this section an experiment that explores the differences in style between the language of government and that of opposition. We provide in Section~\ref{sec:theoretical-motivation} the theoretical background underlying our hypotheses, which we lay out in Section~\ref{sec:hypotheses}. The results of the experiment are detailed and discussed in Section~\ref{sec:vad_differences}.

\subsection{Theoretical motivation}
\label{sec:theoretical-motivation}
Journalists usually prefer to cover politicians with more political power, typically those belonging to the government \citep{doi:10.1177/1464884911427804, wolfsfeld2022making}. Therefore, opposition members tend to act in ways that adhere to what journalists consider newsworthy, in order to attract media attention. For example, opposition members pay more attention to issues raised by the media \citep{zoizner2017how} in order to gauge for societal problems to attack the government \citep{10.1111/j.1460-2466.2006.00005.x}. Moreover, compared to government MPs, opposition members also use more negative and conflictual rhetoric that adheres to what journalists define as newsworthy \citep{Nai_2020, doi:10.1177/13540688231188476}. Since drama and emotional language are key indicators of what makes a story newsworthy \citep{galtung1965structure, harcup2001what}, we expect opposition member to rely more on such language to increase their media appearance.

\subsection{Hypotheses}
\label{sec:hypotheses}
Against this background we posit several hypotheses. We anticipate that the language of the opposition will be more emotionally intense, and we expect the language of coalition members to be more positive compared to the opposition members. 
We also anticipate that government will exhibit more assertive and confident language, since its members are in a position of power and authority and are more likely to use language that reflects their control and leadership.

We thus anticipate that differences in style between coalition and opposition will be reflected in the VAD values of the Knesset data. We expect Valence values to be higher in utterances of coalition members, since the language of government is more positive than that of the opposition. We also expect Arousal values to be higher in the opposition, demonstrating emotionally intense language. Finally, we expect Dominance values to be higher in the coalition, showing assertiveness and confidence. 

To capture these trends we defined several measures, which we computed at the protocol level, for each of the VAD dimensions:
\begin{itemize}
\item Mean VAD values of sentences in the protocol.\footnote{We also experimented with the median, rather than the mean, and the results were similar.}
\item Ratio of high value sentences in the protocol. We defined `high' as values above~0.7.
\item Ratio of low value sentences in the protocol. We defined `low' as values below~0.3. 
\end{itemize}

\subsection{Differences across Emotional Dimensions between Government and Opposition}
\label{sec:vad_differences}
Our goal here is to measure the differences in style, in terms of VAD, between coalition and opposition members, in order to validate these expectations and verify that our VAD approach is applicable, before using it to evaluate polarization. In this task we only considered corpus sentences that are associated with speakers identified as Knesset members, and used the standard t-test to determine the directionality and significance of the differences between two groups: 
We calculated the metrics discussed above \emph{separately} for the coalition and the opposition. The values are computed per protocol; the number of data points is hundreds to thousands, depending on the committee.

Table~\ref{tbl:t_test_coalition_opposition} reports the results.
We test for each committee and for each VAD measure, whether the value is (significantly, with $p{<}0.05$) higher for the coalition or the opposition; results matching our expectations are displayed with grey background.
Evidently, Knesset members of the coalition presented statistically significantly \textit{higher} mean Valence values (\texttt{V\_mean}) in most committees. Additionally, a higher percentage of coalition sentences had high V values. Conversely, opposition members showed a higher percentage of sentences with \textit{lower} V values. These findings indicate that the V values are higher in the coalition compared to the opposition, suggesting a more positive discourse among coalition members, in line with our intuition and theoretical motivation (Section ~\ref{sec:theoretical-motivation}).
Similarly, mean Arousal values (\texttt{A\_mean}) are indeed higher for the opposition compared to the coalition, in most committees, indicating a more emotionally-intense discourse. Additionally, the higher D values of the coalition indicate greater dominance and confidence. Both findings are aligned with our expectations, corroborating our conjecture that the VAD approach is an effective way to compute emotions in Knesset texts.

\begin{table*}[hbt]
\centering
\resizebox{\textwidth}{!}{
\begin{tabular}{lcccccccccc}
\toprule
\textbf{Committee} & \textbf{\texttt{V\_mean}} & \textbf{\texttt{V\_high}} & \textbf{\texttt{V\_low}} & \textbf{\texttt{A\_mean}} & \textbf{\texttt{A\_high}} & \textbf{\texttt{A\_low}} & \textbf{\texttt{D\_mean}} & \textbf{\texttt{D\_high}} & \textbf{\texttt{D\_low}} \\
\midrule
Finance & \cellcolor{gray!30}{gov} & \cellcolor{gray!30}{gov} & \cellcolor{gray!30}{opp} & \cellcolor{gray!30}{opp} & \cellcolor{gray!30}{opp} & \cellcolor{gray!30}{gov} & \cellcolor{gray!30}{gov} & \cellcolor{gray!30}{gov} & \cellcolor{gray!30}{opp} \\
Constitution, Law and Justice & \cellcolor{gray!30}{gov} & \cellcolor{gray!30}{gov} & \cellcolor{gray!30}{opp} & --- & \cellcolor{gray!30}{opp} & --- & \cellcolor{gray!30}{gov} & \cellcolor{gray!30}{gov} & \cellcolor{gray!30}{opp} \\
Economic Affairs & --- & --- & gov & \cellcolor{gray!30}{opp} & --- & \cellcolor{gray!30}{gov} & --- & --- & gov \\
Internal Affairs and Environment & \cellcolor{gray!30}{gov} & \cellcolor{gray!30}{gov} & \cellcolor{gray!30}{opp} & --- & \cellcolor{gray!30}{opp} & --- & \cellcolor{gray!30}{gov} & \cellcolor{gray!30}{gov} & \cellcolor{gray!30}{opp} \\
Labor, Welfare and Health & \cellcolor{gray!30}{gov} & \cellcolor{gray!30}{gov} & \cellcolor{gray!30}{opp} & --- & \cellcolor{gray!30}{opp} & opp & \cellcolor{gray!30}{gov} & \cellcolor{gray!30}{gov} & \cellcolor{gray!30}{opp} \\
Education, Culture and Sports & \cellcolor{gray!30}{gov} & \cellcolor{gray!30}{gov} & \cellcolor{gray!30}{opp} & \cellcolor{gray!30}{opp} & \cellcolor{gray!30}{opp} & \cellcolor{gray!30}{gov} & \cellcolor{gray!30}{gov} & \cellcolor{gray!30}{gov} & \cellcolor{gray!30}{opp} \\
State Control & --- & \cellcolor{gray!30}{gov} & --- & \cellcolor{gray!30}{opp} & --- & \cellcolor{gray!30}{gov} & --- & \cellcolor{gray!30}{gov} & --- \\
Status of Women and Gender Equality & \cellcolor{gray!30}{gov} & \cellcolor{gray!30}{gov} & --- & --- & --- & --- & --- & --- & --- \\
Immigration, Absorption and Diaspora & \cellcolor{gray!30}{gov} & \cellcolor{gray!30}{gov} & \cellcolor{gray!30}{opp} & \cellcolor{gray!30}{opp} & \cellcolor{gray!30}{opp} & \cellcolor{gray!30}{gov} & \cellcolor{gray!30}{gov} & \cellcolor{gray!30}{gov} & \cellcolor{gray!30}{opp} \\
House & \cellcolor{gray!30}{gov} & \cellcolor{gray!30}{gov} & \cellcolor{gray!30}{opp} & \cellcolor{gray!30}{opp} & \cellcolor{gray!30}{opp} & \cellcolor{gray!30}{gov} & \cellcolor{gray!30}{gov} & \cellcolor{gray!30}{gov} & \cellcolor{gray!30}{opp} \\
Rights of the Child & --- & --- & --- & \cellcolor{gray!30}{opp} & --- & \cellcolor{gray!30}{gov} & --- & --- & --- \\
Science and Technology & \cellcolor{gray!30}{gov} & \cellcolor{gray!30}{gov} & --- & \cellcolor{gray!30}{opp} & --- & --- & \cellcolor{gray!30}{gov} & \cellcolor{gray!30}{gov} & --- \\
Public Petitions & \cellcolor{gray!30}{gov} & \cellcolor{gray!30}{gov} & \cellcolor{gray!30}{opp} & \cellcolor{gray!30}{opp} & \cellcolor{gray!30}{opp} & --- & \cellcolor{gray!30}{gov} & \cellcolor{gray!30}{gov} & \cellcolor{gray!30}{opp} \\
\bottomrule
\end{tabular}}
\caption{Coalition vs.\ Opposition in the Knesset data. According to t-test, at significance level $p{<}0.05$: which group has a higher value, coalition (gov) or opposition (opp). Dashes indicate that the difference between the groups is not statistically significant. The values that match our expectations regarding the differences between the two groups are colored with grey background.}
\label{tbl:t_test_coalition_opposition}
\end{table*}

\section{Polarization Trends Over Time}
\label{sec:vad_over_time}


We can now address our main research question, namely the assessment of affective polarization in the Knesset proceedings. We first operationalize polarization in terms of the VAD measures, and then use this approach to measure polarization trends over time in the Knesset data.

\subsection{Operationalization: Characterizing Affective Polarization using VAD}
\label{sec:vad_characterization}

We anticipate that polarized discussions will manifest, in part, by \textit{high values of Arousal}, because high Arousal values indicate the use of emotional, keen, and intensive language, as opposed to a more calm and indifferent language. Such language is more likely to appear in arguments, disagreements, and emotionally polarized discourse, as studies show that anger, disgust, and toxicity scores are substantially higher in interactions between people of opposing ideologies \citep{lerman2024affectivepolarizationdynamicsinformation}. 
Additionally, we anticipate that polarization will be expressed in a \textit{high variance of Valence values}. When one speaks in favor of a certain subject, the Valence values should be high because the statement is positive. On the other hand, statements against a certain topic will be reflected in low Valence values. In a polarized discussion, where speakers disagree with each other, we expect to see many statements from both ends of the scale, resulting in high variance in Valence. In addition, since affective polarization manifests as positive feelings and interactions between people of the same ideology and negative ones towards people of opposing ideologies \citep{https://doi.org/10.1111/pops.12955, lerman2024affectivepolarizationdynamicsinformation}, in a polarized discussion we expect to see both positive and negative Valence values, again resulting in high variance. Therefore, in this work, we operationalize affective polarization as \emph{the combination of high variance in Valence and high values of Arousal}. We expect to discover an increase in these measures over the years.

\subsection{Results}
\label{sec:metrics_computation_trend_analysis}

We sorted the protocols of each committee in chronological order so that each protocol established a point in time. We associated VAD values with the sentences of each protocol, and calculated for each protocol its VAD measures as described above. 
To examine trends over time, we applied the \textit{Mann-Kendall test} \citep{Mann1945NonparametricTA, kendall1948rank}, which statistically assesses if there is a monotonic upward or downward trend of a variable of interest in time series data, to the series of values of each measure. 

\begin{figure}[hbt]
\centering
\includegraphics[width=0.7\columnwidth]{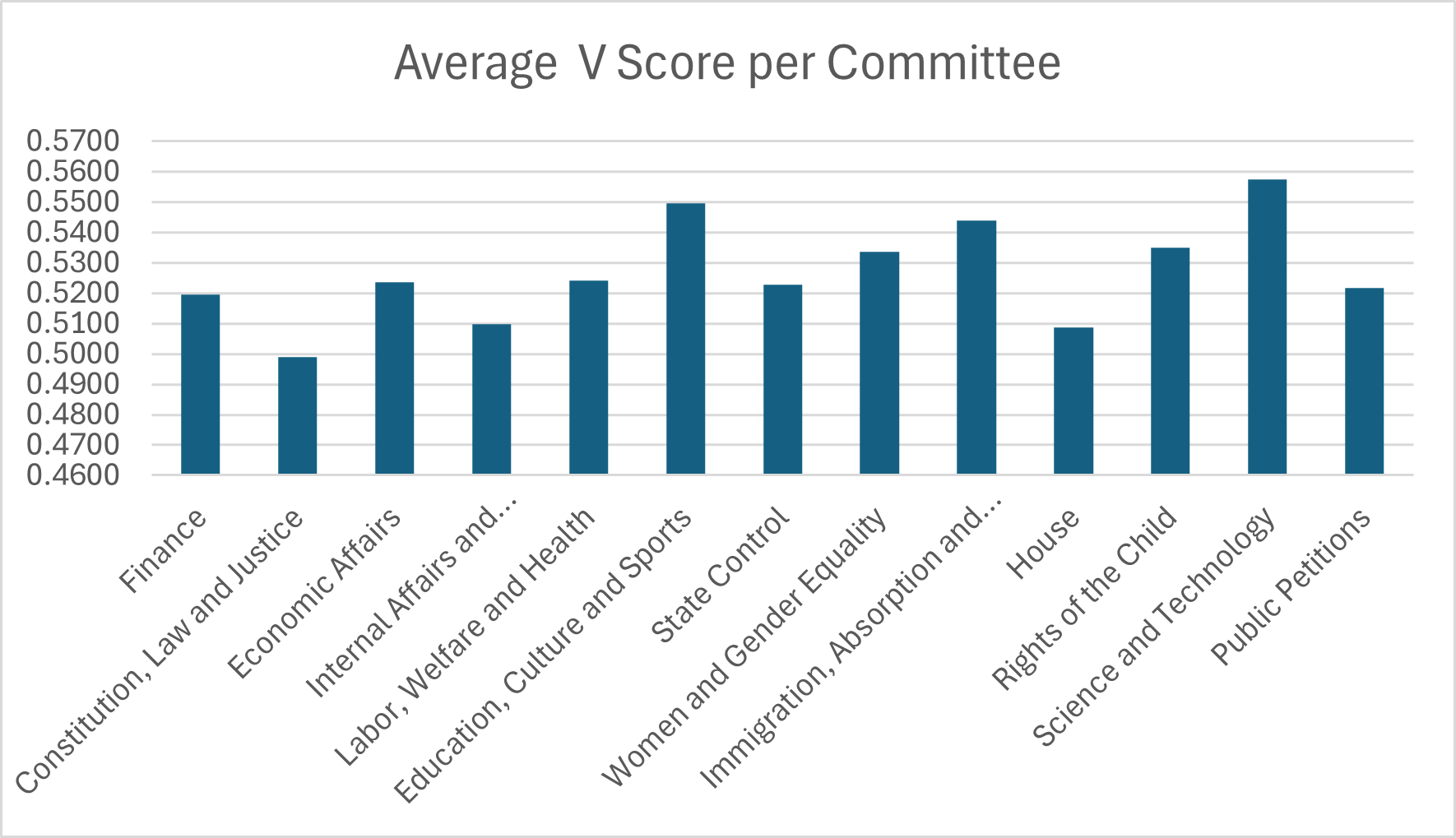}

\vspace{1em}
\includegraphics[width=0.7\columnwidth]{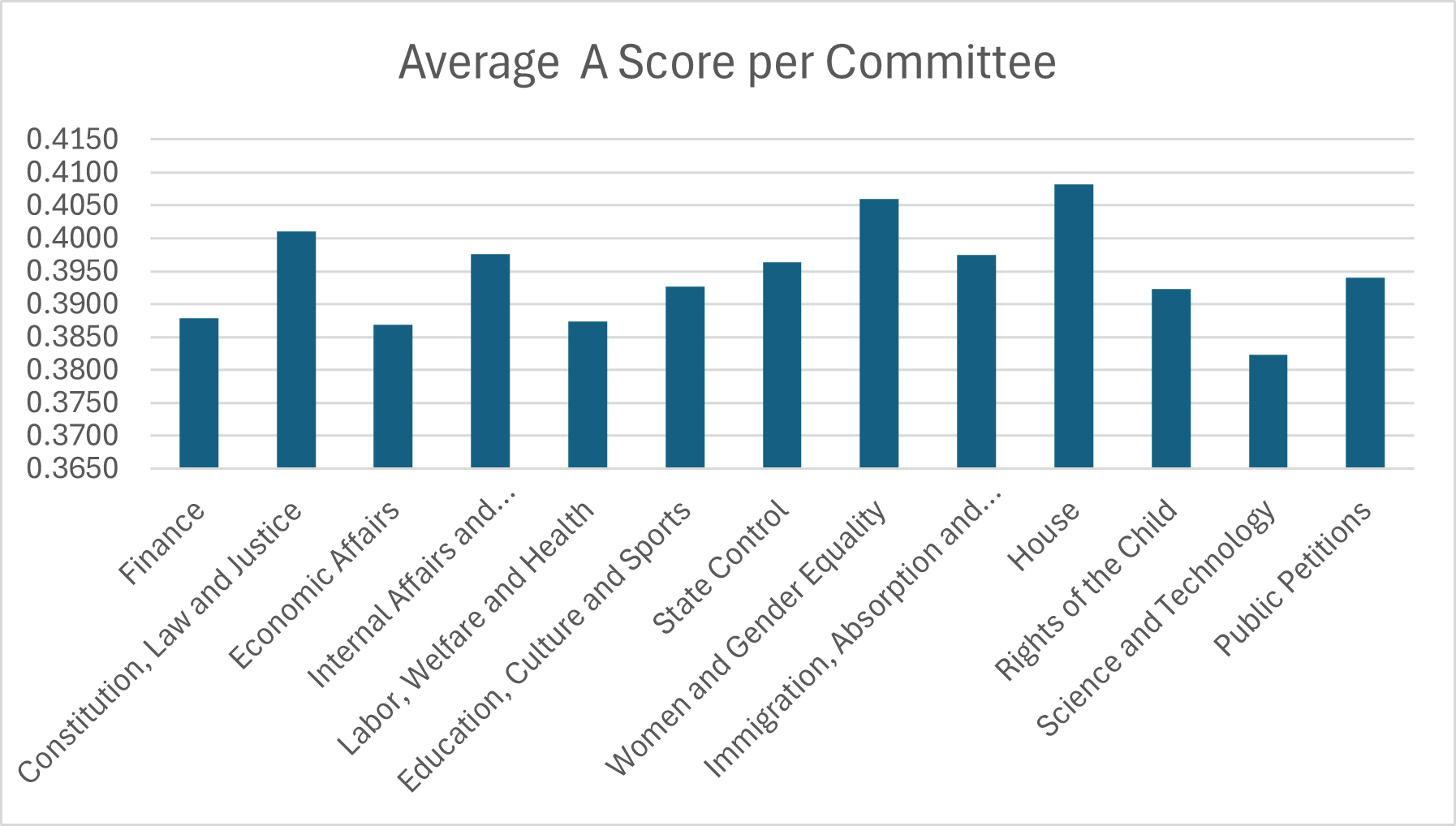}

\vspace{1em}
\includegraphics[width=0.7\columnwidth]{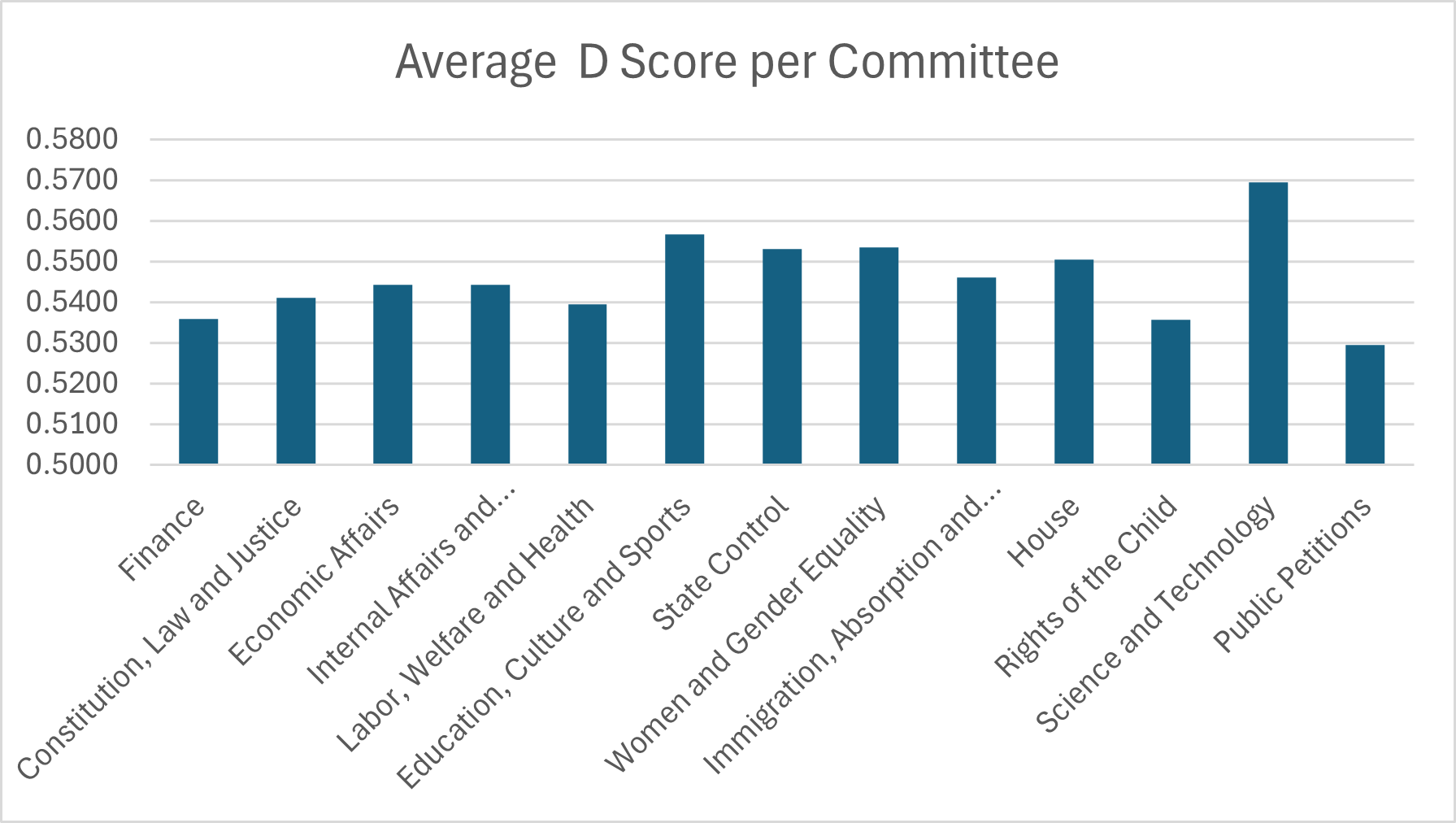}
\caption{Average Valence, Arousal and Dominance values over all sentences in each of the committees.}
\label{fig:avg_avd_per_committee}   
\end{figure}

Figure~\ref{fig:avg_avd_per_committee} depicts the average V, A and D scores for each committee, respectively. 
It shows, for example, higher Valence values in committees such as Science and Technology or Education, Culture and Sports, as one could expect; higher Arousal values in the House committee, which reflects emotionally-loaded discussions; and lower Arousal values in, e.g., the Science and Technology committee, reflecting less polarizing discussions there.
More visualizations are given in Appendix~\ref{app:corpus_statistics}.

\begin{figure*}[hbt]
\centering
\includegraphics[width=0.48\textwidth]{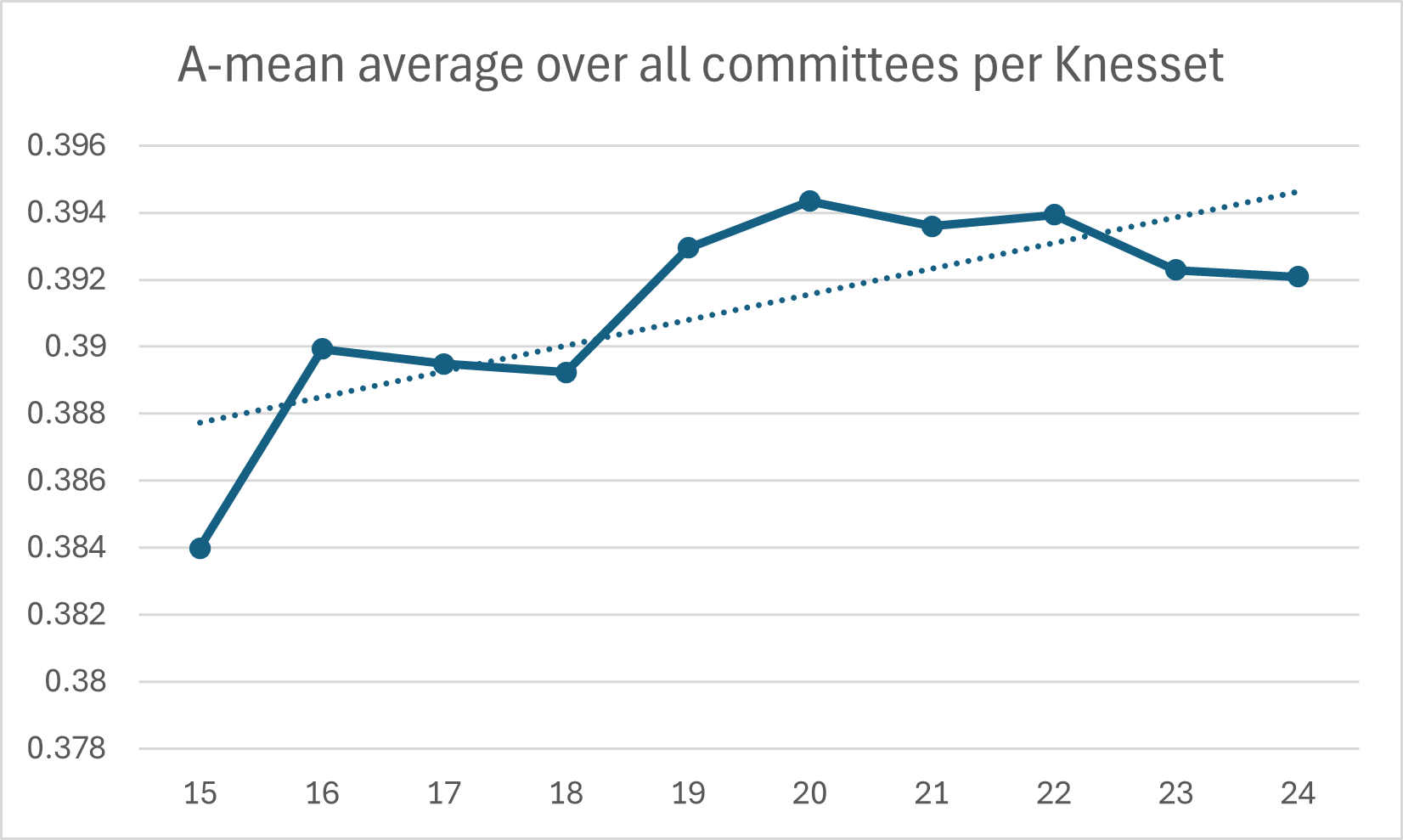}
\includegraphics[width=0.48\textwidth]{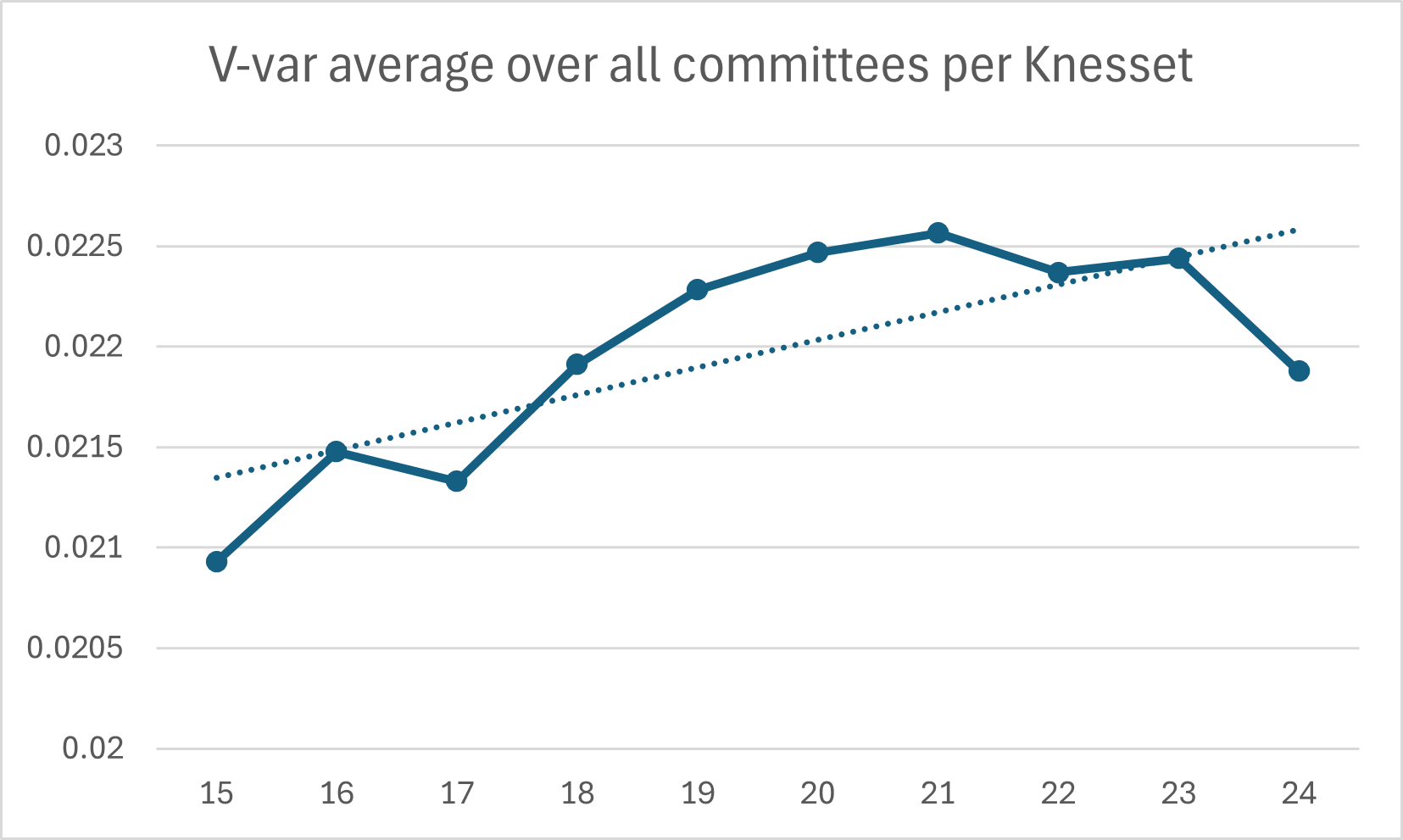}
\caption{Average A-mean and V-var over all committees, over time (Knesset session). Dashed lines represent the automatically computed trend lines. }
\label{fig:a-mean-v-var-avg}   
\end{figure*}

Figure~\ref{fig:a-mean-v-var-avg} shows the values of A-mean and V-var, averaged over all Knesset committees, on a time scale that shows one data point for each Knesset session, from the 15th Knesset to the 24th.\footnote{Whenever the Israeli parliament is elected, a new Knesset \emph{session} begins. Legally, the length of a session is four years, but many sessions ended prematurely. Our committee data reflects protocols from the 15th Knesset up until the (current) 25th, but we excluded the 25th Knesset from these graphs because it is still in session.}
The upward trend of these values, as a function of time, is clearly evident.

The full results, broken down by committee, are presented in Table~\ref{tbl:overtime_trends_mann_kendall}. They show for each VAD measure whether its series of values over time shows a significant increasing or decreasing trend, as determined by the Mann-Kendall test. For completeness, the table includes all the measures for all three VAD dimensions. Colored with grey background are the entries that matched our expectations regarding the variance of Valence and the Arousal values. 

As we expected, in most committees there is a statistically significant increase in the variance of Valence (\texttt{V\_var}), indicating disagreements in the conversations. The results also show an increase in the values of Arousal (\texttt{A\_mean}), indicating a rise in the level of intensity in the discourse, and a decrease in the percentage of low-Arousal sentences in most committees, supporting this claim. Based on our operationalization of polarization, as defined above, these results demonstrate a significant and reliably-detected trend of increasing affective polarization in the Knesset data over the years.

\begin{table*}[hbt]
\centering
\resizebox{\textwidth}{!}{
\begin{tabular}{lccccccccccc}
\toprule
\textbf{Committee} & \texttt{V\_mean} & \textbf{\texttt{V\_var}} & \texttt{V\_high} & \texttt{V\_low} & \textbf{\texttt{A\_mean}} & \textbf{\texttt{A\_high}} & \textbf{\texttt{A\_low}} & \texttt{D\_mean} & \texttt{D\_high} & \texttt{D\_low} \\
\midrule
Finance & $\downarrow$ & \cellcolor{gray!30}{↑} & --- & ↑ & \cellcolor{gray!30}{↑} & \cellcolor{gray!30}{↑} & \cellcolor{gray!30}{↓} & ↓ & ↓ & ↑ \\
Constitution, Law and Justice & ↑ & \cellcolor{gray!30}{↑} & ↑ & ↓ & --- & \cellcolor{gray!30}{↑} & --- & ↓ & ↓ & ↑ \\
Economic Affairs & --- & \cellcolor{gray!30}{↑} & ↑ & ↑ & \cellcolor{gray!30}{↑} & \cellcolor{gray!30}{↑} & \cellcolor{gray!30}{↓} & ↓ & ↓ & ↑ \\
Internal Affairs and Environment & ↑ & \cellcolor{gray!30}{↑} & ↑ & ↓ & \cellcolor{gray!30}{↑} & \cellcolor{gray!30}{↑} & \cellcolor{gray!30}{↓} & ↓ & ↓ & ↑ \\
Labor, Welfare and Health & --- & \cellcolor{gray!30}{↑} & ↑ & --- & \cellcolor{gray!30}{↑} & \cellcolor{gray!30}{↑} & \cellcolor{gray!30}{↓} & ↓ & ↓ & ↑ \\
Education, Culture and Sports & ↑ & \cellcolor{gray!30}{↑} & ↑ & ↓ & \cellcolor{gray!30}{↑} & \cellcolor{gray!30}{↑} & \cellcolor{gray!30}{↓} & ↓ & ↓ & ↑ \\
State Control & ↑ & \cellcolor{gray!30}{↑} & ↑ & --- & \cellcolor{gray!30}{↑} & \cellcolor{gray!30}{↑} & \cellcolor{gray!30}{↓} & ↓ & ↓ & ↑ \\
Women and Gender Equality & --- & \cellcolor{gray!30}{↑} & ↑ & --- & \cellcolor{gray!30}{↑} & \cellcolor{gray!30}{↑} & \cellcolor{gray!30}{↓} & ↓ & ↓ & ↑ \\
Immigration, Absorption and Diaspora & ↑ & ↓ & ↑ & ↓ & \cellcolor{gray!30}{↑} & \cellcolor{gray!30}{↑} & \cellcolor{gray!30}{↓} & --- & ↓ & --- \\
House & --- & \cellcolor{gray!30}{↑} & ↑ & ↑ & \cellcolor{gray!30}{↑} & \cellcolor{gray!30}{↑} & \cellcolor{gray!30}{↓} & ↓ & ↓ & ↑ \\
Rights of the Child & ↑ & \cellcolor{gray!30}{↑} & $\uparrow$ & --- & \cellcolor{gray!30}{↑} & \cellcolor{gray!30}{↑} & \cellcolor{gray!30}{↓} & --- & --- & ↑ \\
Science and Technology & --- & --- & ↑ & --- & \cellcolor{gray!30}{↑} & \cellcolor{gray!30}{↑} & \cellcolor{gray!30}{↓} & ↓ & ↓ & --- \\
Public Petitions & ↑ & --- & ↑ & ↓ & \cellcolor{gray!30}{↑} & \cellcolor{gray!30}{↑} & \cellcolor{gray!30}{↓} & ↓ & ↓ & --- \\
\bottomrule
\end{tabular}}
\caption{Polarization over time in the Knesset data. Results at significance level $p{<}0.05$ are indicated by arrows: ↑ increasing or ↓ decreasing trends. Dashes indicate non-significant trends. The trends that match our expectations regarding our polarization indicators (\texttt{A\_(*)} and \texttt{V\_var}) are depicted with grey background.}
\label{tbl:overtime_trends_mann_kendall}
\end{table*}

Another interesting trend that emerges from Table~\ref{tbl:overtime_trends_mann_kendall} is a decline in D values over the years. Dominance measures the degree of control or influence a speaker or statement exhibits. This decline may reflect a shift toward more ambiguous and less authoritative speech, in which politicians avoid making clear commitments or decisive statements. We also analyzed the trends of D values over time separately for the coalition and the opposition, and  discovered that the decline is consistent in both groups. Although the Dominance values are higher in the coalition, as expected, the decline over time is evident across political divides. This suggests that while polarization may be increasing—-leading to more heated disagreements and stronger partisan rhetoric-—politicians are simultaneously becoming less explicit in taking responsibility or making decisive statements, with a growing tendency to avoid firm commitments, even among those in control and in leading positions.
Similarly, the data in Table~\ref{tbl:overtime_trends_mann_kendall} suggest an overall increase in the values of Valence, in half of the committees.
We leave a more detailed analysis of these trends to future research.

\subsection{Accounting for (some) Confounding Factors}
Encouraged by the trends in \texttt{V\_var} and \texttt{A\_mean} found in Section~\ref{sec:vad_over_time}, we now turn to a more fine-grained investigation of the polarization phenomenon, accounting also for factors that may affect emotional dimensions, introducing potential confounds into the analysis. As a concrete example, high variance in Valence can be caused by a diverse committee, with balanced representation of both government and opposition. We believe that the more heterogeneous a committee meeting is, including MPs from both sides of the political map (represented here as coalition vs.\ opposition), the more polarized the discourse would be. In contrast, the more homogeneous a meeting is, the less polarized we expect the discourse to be. Additionally, it has been shown that men and women differ in VAD attributes of their spontaneous language \citep[e.g., on social media,][]{DBLP:journals/corr/abs-2008-05713}, and that there are difference in the language style between men and women in the Knesset proceedings \citep{Goldin2024TheKC}, so gender representation in a committee may contribute to the observed VAD fluctuations. We confirm and extend our findings in Section~\ref{sec:vad_over_time} by casting the diachronic usecase as a linear regression analysis, explicitly modeling the temporal axis, as well as several additional confounds, as predictor (independent) variables. Specifically, we define three predictors: a point at time (\texttt{TP} in the [1..$n$] range, where each protocol represents a sequential point in time and $n$ is the number of protocols), the ratio of government members in the committee (\texttt{RatioG}), and the ratio of female members in the committee (\texttt{RatioF}). We build two regression models, predicting \texttt{V\_var} and \texttt{A\_mean}, the two outcome (dependent) variables, in the committee protocol. 

The results reported in Table~\ref{tbl:overtime_trends_mann_kendall} imply that VAD values may vary across the committees -- not a surprising finding, given that some topics are more likely to trigger emotional language than others. We model the categorical committee predictor (\texttt{Comm}) as a \emph{fixed effect} in our analysis, facilitating \emph{interactions} between the committee and other predictors in our unified model. A positive and significant coefficient of the time point (\texttt{TP}) predictor will be indicative of the correlated growth in the dependent variable over time; that is, increasing polarization across emotional dimensions.
Formally, a predictive model with interactions is defined as (\texttt{predicted\_var} stands for the predicted variable, namely \texttt{A\_mean} or \texttt{V\_var} in our case):
\begin{equation}
\texttt{predicted\_var} \sim \texttt{Comm*TP + Comm*RatioC + Comm*RatioF}
\label{eq:fixed_effects}
\end{equation}

\noindent
where `*' denotes the first- and second-order effects of two variables.  
Practically, while the model in this equation is the most common way of modeling fixed effects for \emph{predictive analysis}, its interpretation is not intuitive. An arbitrary category (also denoted as `base' category) is eliminated to avoid multi-collinearity, and the contribution of various predictors as mirrored by their coefficients is reported as a relative change (positive or negative) to the base category coefficient. Similar interpretation is used for statistical indicators: $p$-values and confidence intervals (CIs). Since we are only interested in comparative analysis of the coefficients (reflecting their contribution to the predicted variables \texttt{A\_mean} and \texttt{V\_var}), we adopt a more easily interpretable variant of the formulation in Equation~\ref{eq:fixed_effects}, defined as follows:
\begin{equation}
\texttt{predicted\_var} \sim \texttt{Comm:TP + Comm:RatioC + Comm:RatioF}
\label{eq:fixed_effects_split}
\end{equation}

\noindent
where `:' (instead of `*') introduces a key difference -- it is equivalent to fitting a separate regression model for each individual \texttt{Comm} (committee) category, eliminating the need in removing a base category variable.\footnote{We normally use explicit interactions (`*') with the omitted level for \textit{modeling purposes}. Note that the model's goodness of fit ($r$-squared) is identical in both equations.} The adjusted model allows for direct interpretation of per-committee variable contribution (sign, value, significance): time point (\texttt{TP}), the ratio of government and female members: \texttt{RatioC} and \texttt{RatioF}, respectively. Recall that we are specifically interested in polarization evidence potentially introduced by the \texttt{TP} variable: its \textit{positive} and \textit{significant} coefficients will corroborate the hypothesis of affective polarization in the  committee proceedings language.

Tables~\ref{tbl:ols_v_var} and~\ref{tbl:ols_a_mean} report example findings for \texttt{V\_var} and \texttt{A\_mean} in the proceedings of the \emph{Economic Affairs} committee; the values are very similar across most committees (full results are reported in Appendix~\ref{app:full_ols}). Evidently, the time point measure is statistically significant and the corresponding coefficients are positive. This indicates an increase in the predicted polarization variables as a function of time. As shown in Appendix~\ref{app:full_ols}, these results are similar in all committees  for the \texttt{V\_var} measure and in 11 out of 13 committees in the \texttt{A\_mean} measure, where the only committees that did not exhibit a significantly positive rise over time for this measure are the `Status of Women and Gender Equality' and the `Constitution, Law and Justice' committees. Therefore, even when isolated from other potential confounds, these results show that polarization increases over time, reconfirming our hypothesis. 

Additionally, the results show that the proportion of coalition members is a significant predictor in both models (top and bottom) and the corresponding coefficients are negative, as expected. As shown in Appendix~\ref{app:full_ols}, these results are similar in 8 out of 13 committees for both measures, indicating that in most committees, more homogeneous meetings tend to have lower \texttt{V\_var} and \texttt{A\_mean} values, suggesting less polarization. This confirms our additional conjecture that the composition of the meeting influences the level of polarization in its discourse. 
The results regarding the proportion of female participants in the committees did not generally reveal a clear trend: In many cases the values were not statistically significant, and when they were, they were mixed—some positive and some negative—indicating that the proportion of women in the committees does not significantly impact the polarization of the discourse. Note that in the two committees reflected in Tables~\ref{tbl:ols_v_var} and~\ref{tbl:ols_a_mean} the effect is significant and positive, but see Appendix~\ref{app:full_ols} for the full details.

\begin{table}[hbt]
\centering
\begin{tabular}{lrrrr}
\toprule
\textbf{Variable} & $\beta{\times}100$  &  P>|t| &   [0.025 &   0.975] \\
\midrule
\texttt{const} &     38.1 &   0.000 & 0.380   & 0.382     \\
\texttt{TP} &        0.30 &    0.000 & 0.002   & 0.004     \\
\texttt{RatioF} &    0.39 &    0.000 & 0.003   & 0.005     \\
\texttt{RatioG} &    -0.23 &   0.000 & -0.003  & -0.002    \\
\hline
\end{tabular}
\caption{OLS regression results for the Economic Affairs committee: predicted variable \texttt{V\_var}. We report $p$-values and confidence intervals at the 95\% confidence level.}
\label{tbl:ols_v_var}
\end{table}

\begin{table}[hbt]
\centering
\begin{tabular}{lrrrr}
\toprule
\textbf{Variable} & $\beta{\times}100$  &  P>|t| &   [0.025 &   0.975] \\
\midrule
\texttt{const} &     2.13 &    0.000 &    0.021 &  0.021       \\
\texttt{TP} &        0.04 &    0.000 &    0.000 &  0.001       \\
\texttt{RatioF} &    0.04 &    0.000 &    0.000 &  0.001       \\
\texttt{RatioG} &    -0.02 &   0.000 &   -0.000 &  -8.7e-05    \\
\hline
\end{tabular}
\caption{OLS regression results for the Economic Affairs committee: predicted variable \texttt{A\_mean}. We report $p$-values and confidence intervals at the 95\% confidence level.}
\label{tbl:ols_a_mean}
\end{table}

\section{Conclusion and Future work}
We presented a novel method for measuring affective polarization in parliamentary proceedings, based on stylistic analysis that assesses emotional language across three dimensions. Applying this method to the protocols of the Knesset, we showed a statistically significant trend of increasing polarization over time. We also showed that the emotional styles of government and opposition differ. 

These results can be extended in various ways. In particular, we are interested in understanding \emph{why} polarization increases: is it a bottom-up trend (i.e., citizens are more polarized, including on social media, which affect politicians’ discourse), or a top-down process (i.e., politicians are polarized which makes the public polarized).
In the future, we plan to compare the trends we revealed in the Knesset data to those found on general media in order to determine temporal causality. 

We found a consistent decline in Dominance values over time for both the coalition and the opposition, suggesting a growing tendency among politicians to avoid firm commitments. Future research could further investigate the linguistic or rhetorical strategies underlying this trend. 

While this study focused on committee protocols due to their spontaneous nature and clear topic divisions, future work will extend the analysis also to plenary protocols. This will allow us to examine whether polarization is also present in structured and prepared discourse. 
On a more technical level, we also plan to further evaluate our fine-tuning technique on other models, datasets and tasks in order to asses its generalizability and robustness for broader use.

Finally, because our main dataset is fully compatible with a host of parliamentary proceedings corpora for dozens of languages, and because our methodologies are straight-forward and all the resources and code used in this work are freely available, we trust that our findings can relatively easily be adapted to various other languages, and we look forward to exploring avenues for collaborating with researchers worldwide on extending our results to other languages, cultures and domains.

\section*{Limitations}
In this paper we attribute the rise of emotional discourse in the Knesset to polarization. There could possibly be other explanations, such as the increasing dominance of media logic in politics over the years \citep{doi:10.1177/1940161208319097}: politicians gradually adopt rhetorical styles that attract journalists, and emotional language is one important element that journalist want.
Other confounding factors could be inconsistent editing of the protocols over time (we suspect that the earlier protocols were more significantly edited, whereas recent ones more accurately reflect the actual discussions), limited and/or biased distribution of topics in some committees, and more. 
The issues discussed in any particular committee may also have changed over time, which may also add noise to our analyses.

A technical limitation stems from the fact that we train our regression models mainly on words, with very few additional annotated sentences, while we use them to predict VAD values of full sentences. Unfortunately, there are no additional resources for training VAD models in Hebrew, to the best of our knowledge, but this could (and probably does) affect the quality of our prediction, as reflected by relatively low Pearson correlation values, especially for~A and~D. \cg{Although these correlations were lower, they remained robust (above 0.5). Importantly, the inter-annotator agreement for arousal and dominance was also comparatively lower, indicating that these dimensions are more subjective than Valence. Therefore, it is reasonable to expect that models would find these values more challenging to predict.}

\section*{Ethical Considerations}
The main dataset we used in this work is the Knesset Corpus, which is publicly-available and adheres to standard ethical measures. In particular, the raw data and much of the meta-data included in this corpus are distributed by the Knesset itself. While the methodology we introduced in this work can be used to assign emotional values to natural language texts, we cannot foresee potential for abuse or dual use of these methods, at least no more than any common methods for sentiment analysis.

We employed three linguists (two women, one man, all native Hebrew speakers residing in Israel) to translate, fix and enrich the Hebrew VAD lexicon and to annotate the sampled Knesset sentences for VAD. The annotators were recruited directly, not through a crowd-sourcing platform, and received an hourly payment that was approximately 2.5 times higher than the minimum wage. No human participants were required for this project.

No AI assistants were used in this project or in the write-up of this paper.

\begin{acknowledgments}
We thank the three anonymous reviewers and the CL editors for their constructive and useful comments. The final version of this paper is much improved due to their feedback and support.
We are extremely grateful to Alon Zoizner for unwavering support and fruitful discussions. 
We thank Rotem Dror for her helpful consultation.
We are grateful to Shira Wigderson, Israel Landau and Avia Vaknin for their meticulous annotation efforts.
We thank the Idit PhD Fellowship Program at the University of Haifa for supporting the first author. 
This research was supported by the Ministry of Science \& Technology, Israel under grant no.~3-17990. 

\end{acknowledgments}

\bibliographystyle{compling}
\bibliography{anthology,other}

\appendix
\section{Training Procedure of the \texttt{Knesset-multi-e5-large} model} 
\label{app:finetune_training}
We fine-tuned the \texttt{multilingual-e5-large} model \citep{wang2024multilinguale5textembeddings} from Hugging Face's Transformers library. We used its corresponding \texttt{AutoTokenizer} from Hugging Face, and the model was loaded using the `AutoModelForMaskedLM' class to retrieve the relevant component of the network for training on the MLM task. Some of the weights in the model were randomly initialized since the original model was not pretrained on this task before. We ran the training on a \href{https://slurm.schedmd.com/documentation.html}{SLURM} (Simple Linux Utility for Resource Management) environment. We utilized a distributed training setup with the \href{https://developer.nvidia.com/nccl}{NCCL} backend to leverage multiple GPUs, to ensure efficient parallel processing and gradient synchronization. The Knesset corpus \citep{Goldin2024TheKC}, containing over 32 million sentences, was preprocessed to create text shards for efficient loading and processing. All the texts were preprocessed according to the multilingual model's requirements. Each text entry was formatted and tokenized. We chunked the data into fixed length chunks of 256 tokens. Padding was applied as necessary to ensure fixed lengths. 

The data were split into a training set of 80\% and validation and test sets of 10\% each. We used the following training configuration: The batch size was set to~8. In order to effectively double the batch size without increasing memory usage, we set the gradient accumulation steps to~4. The learning rate was set to $1{\times}10^{-4}$ for most layers which only needed fine-tuning and $1{\times}10^{-6}$ for the layers that were untrained before. We used a weight decay of~0.01 to regularize the model. Early stopping was implemented with a threshold of 0.01 to prevent overfitting. The model was trained for a single epoch. We chose not to continue training beyond this point to avoid potential weight mismatches when ``re-connecting'' the sentence transformer component of the network.

\section{Training Procedure of the \texttt{Regression-Head} model}
\label{app:vad_regression_head_training}
We trained three separate regression models for predicting VAD values using transformer-based sentence encoders with an added regression head. We applied the same procedure to the base encoders of the following models: \emph{DictaBert} \citep{shmidman2023dictabertstateoftheartbertsuite}, Knesset-\emph{DictaBERT} \citep{goldin2024knessetdictaberthebrewlanguagemodel}, \emph{multilingual-e5-large} \citep{wang2024multilinguale5textembeddings}, and the \texttt{Knesset-multi-e5-large} model as described in Section~\ref{sec:multi_knesset_model}.  
Each model’s corresponding \texttt{AutoTokenizer} was used, with preprocessing adapted to the specific architecture. All sentences were tokenized to a maximum length of 128 tokens, truncated when necessary, and padded to a fixed length.  
A single linear regression head with one output unit was appended to the sentence encoder to perform scalar regression. The regression head parameters were initialized using PyTorch’s default linear layer initialization. To preserve pretrained semantic representations, all transformer and pooling layers in the backbone were frozen, and only the regression head parameters were updated during training.  

The training data combined the three datasets detailed in Section~\ref{sec:data_preprocessing}.
Datasets were processed using Hugging Face's \texttt{datasets} and split into training and validation subsets (80\%/20\%). For $k$-fold experiments, the Knesset set was partitioned into five folds with non-overlapping test portions. The training loop used Hugging Face’s \texttt{Trainer} API with \texttt{TrainingArguments} configured as follows: learning rate $2{\times}10^{-5}$, batch size of 16 for training and 32 for evaluation, weight decay $0.01$, and training for three epochs. All training hyperparameters were selected after a series of trial-and-error experiments aimed at identifying the configuration that yielded the best validation performance.

The evaluation metric was Pearson’s correlation coefficient between predictions and gold labels, computed using \href{https://scipy.org/}{SciPy} library, with model selection based on the best validation Pearson score.  
All training was performed in a \href{https://slurm.schedmd.com/documentation.html}{SLURM}-managed multi-GPU environment, as described in Appendix~\ref{app:finetune_training}. Models and tokenizers were saved after training. The trained models were later evaluated on held-out Knesset sentences to compute Pearson correlation and statistical significance values. The results are presented in Table~\ref{tbl:pearson_vad_results_regression_head}.

\section{Evaluation of Off-The-Shelf Models on the Downstream Task of Predicting VAD Values}
\label{app:off-the-shelf-eval-res}
To identify the most suitable LLM model for the downstream task of extracting sentence embeddings and predicting VAD values on the Knesset sentences, we experimented with various off-the-shelf models including \emph{DictaBert} \citep{shmidman2023dictabertstateoftheartbertsuite}, \emph{sentence-transformers-alephbert} \citep{seker2021alephberta}, \emph{alephbertgimmel-base-512} \citep{gueta2023largepretrainedmodelsextralarge}, Knesset-\emph{DictaBERT} \citep{goldin2024knessetdictaberthebrewlanguagemodel}, \emph{multilingual-e5-large} \citep{wang2024multilinguale5textembeddings}, and others. We assessed their performance on the downstream task of predicting VAD scores using the binomial regression models described in Section~\ref{sec:vad_binom_models}. We used Pearson correlation coefficient for evaluation, chosen for its effectiveness in measuring the linear correlation between variables. 

For initial elimination, we conducted a simple experiment to compare the models: we used only the Hebrew VAD lexicon for both training and testing, splitting the data into 80\% training, 10\% validation and 10\% test sets. The results on the test set are presented in Table~\ref{table:pearson-vad-all-models}. While all results were relatively close, the multilingual model yielded the best results for~A and~D, competitive results for~V, and the highest average performance overall. The \emph{sentence-transformer-alephbert} and \emph{alephBertGimmel} model performed worst and were therefore discarded.

\begin{table}[htbp]
\centering
\begin{tabular}{lcccccc}
\toprule
Model  & V & A & D & Average\\
\midrule
sentence-transformer-alephbert & 0.65 & 0.52 & 0.59 & 0.59\\
multilingual-e5-large & 0.66 & \textbf{0.59} & \textbf{0.62} & \textbf{0.62} \\
alephBertGimmel & \textbf{0.67} & 0.52 & 0.59 & 0.59 \\
dictaBERT & \textbf{0.67} & 0.55 & 0.59 & 0.60 \\
Knesset-dictaBERT & 0.65 & 0.56 & 0.58 & 0.60 \\
\bottomrule
\end{tabular}
\caption{Pearson correlation results 
for VAD prediction, trained on the VAD lexicon train set and evaluated on the VAD lexicon test set. Best results in each column are in boldface. }
\label{table:pearson-vad-all-models}
\end{table}

\section{Relating VAD to broader affective frameworks}
\label{app:other-emotional-measures}
In addition to VAD, lexicons of other emotions have been published for various languages, mainly based on the eight prototypical emotions defined by \citet{plutchik1980general}: joy, sadness, anger, fear, disgust, trust, anticipation and surprise \citep{mohammad-turney-2010-emotions,mohammad-yang-2011-tracking,VAD-norms,Mohammad13,Glasgow-norms}.
To investigate how affective polarization, as demonstrated by VAD values, aligns with other emotion-based frameworks, we also explored the NRC Emotion Lexicon \citep{mohammad-etal-2013-nrc}, whose entries are fully contained within the Hebrew VAD lexicons we have validated (Section~\ref{sec:vad_lexicon}).

Following common practices of working with the NRC Emotion Lexicon, and due to the near absence of pre-trained sentence-level models for each NRC dimension, we experimented with word-count techniques. 
We counted the number of words related to each emotion across all the protocols of each committee, normalized by the number of words in each protocol. We sorted the protocols of each committee in chronological order so that each protocol established a point in time and we applied the \textit{Mann-Kendall test} \citep{Mann1945NonparametricTA, kendall1948rank} test to evaluate trends over time for each emotion in every committee.

We found significant time-related upward trends in \emph{joy} and \emph{sadness}, lending support to our V-variance construct. In addition, emotions strongly tied to high arousal, such as \emph{anticipation} and \emph{surprise}, increased with time in nearly all committees, reinforcing our characterization of affective polarization as elevated arousal levels. 
However, roughly one third of committee-emotion combinations did not show any statistically significant trends, and \emph{anger}, which is a high-arousal emotion, showed no significant trend in half of committees, with mixed directions elsewhere.
These mixed outcomes convinced us that lexicon-based counts of discrete emotions, while informative, lack the robustness needed for definitive conclusions. Accordingly, we focused instead on regression models that predict continuous VAD values for our main analyses.

In addition, we examined \emph{HebEmo} \citep{chriqui2021hebert}, 
an off-the-shelf model for emotion classification in Hebrew. We evaluated its robustness on Knesset
sentences by manually assessing its classifications of our 120 Knesset
sentences annotated dataset. The results indicated that the model did not perform satisfactorily
on the Knesset data; it is a binary model that only indicates whether an emotion is or is not present in a sentence,  and a great number of sentences in our dataset that should have been classified as positive
for a certain emotion were assigned a negative prediction. Given that the HebEmo model was trained on user-generated comments on news articles related to Covid-19, a totally different genre from our parliamentary data, the weak generalization power of the model is unsurprising.

\section{Visualization of VAD Statistics in the Corpus}
\label{app:corpus_statistics}

\begin{figure}[hbt]
\centering
\includegraphics[width=0.70\columnwidth]{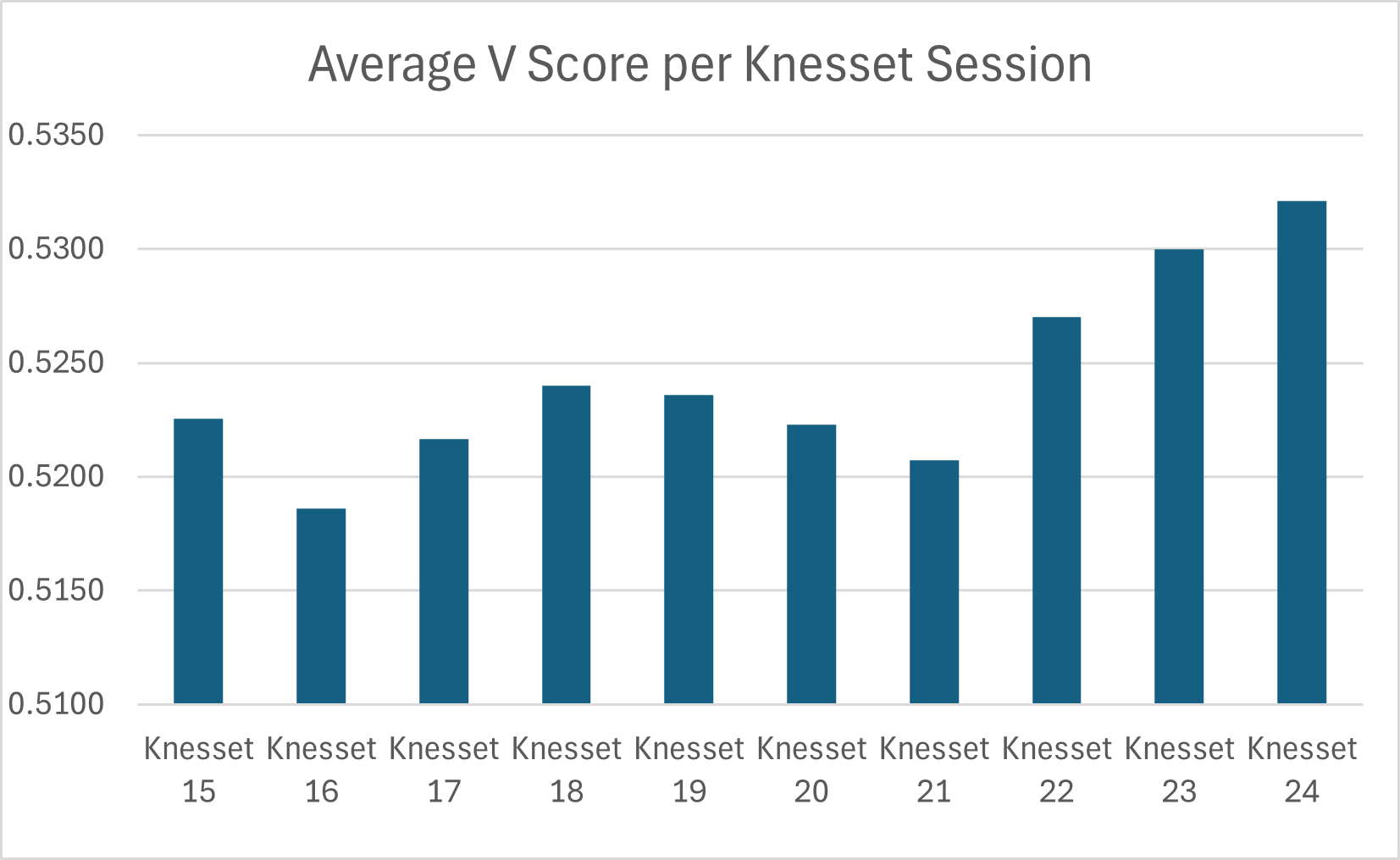}
\caption{Average Valence score over all sentences in each Knesset session.}
\label{fig:avg_v_per_knesset}   
\end{figure}

\begin{figure}[hbt]
\centering
\includegraphics[width=0.70\columnwidth]{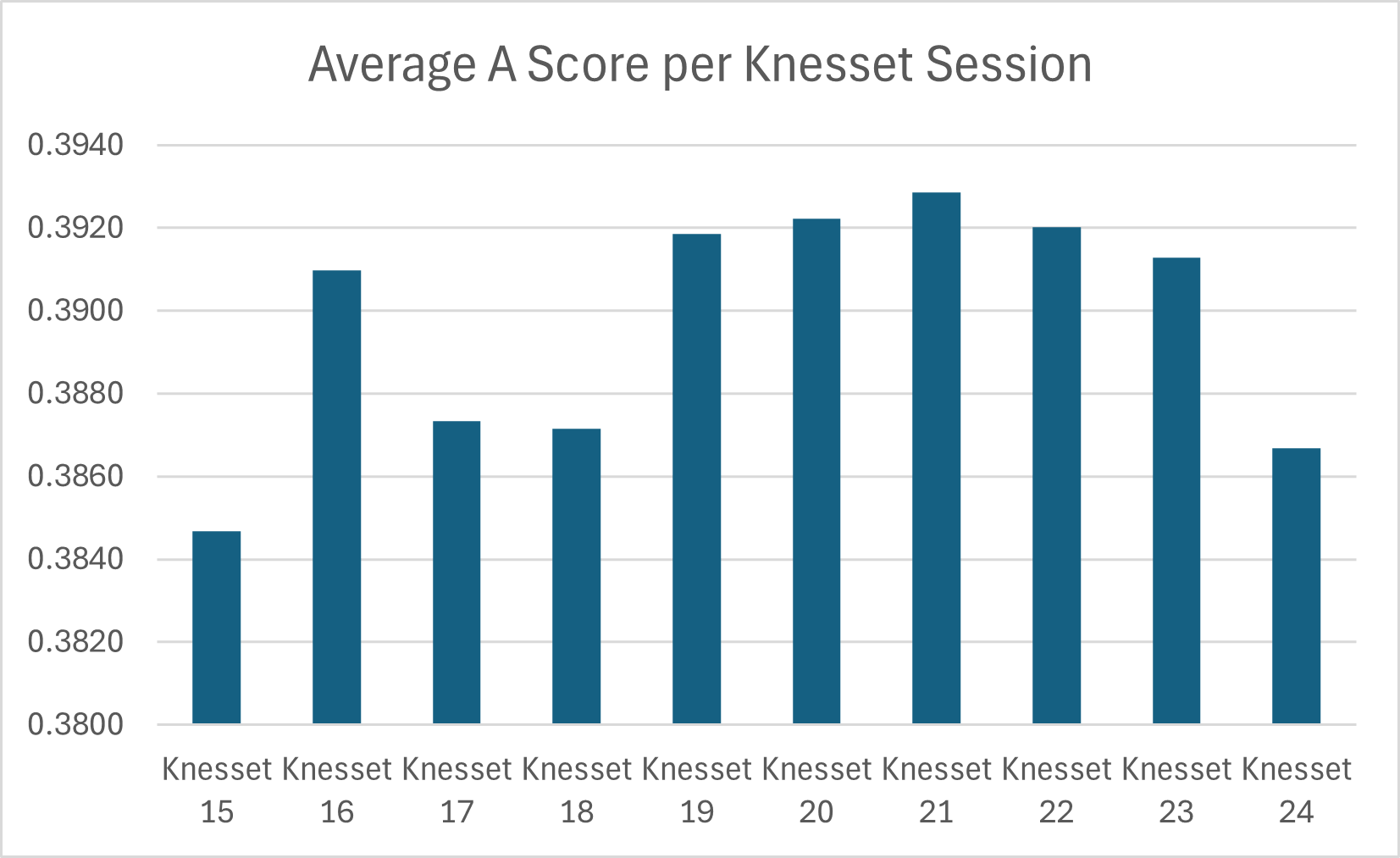}
\caption{Average Arousal score over all sentences in each Knesset session.}
\label{fig:avg_a_per_knesset}   
\end{figure}

\begin{figure}[hbt]
\centering
\includegraphics[width=0.70\columnwidth]{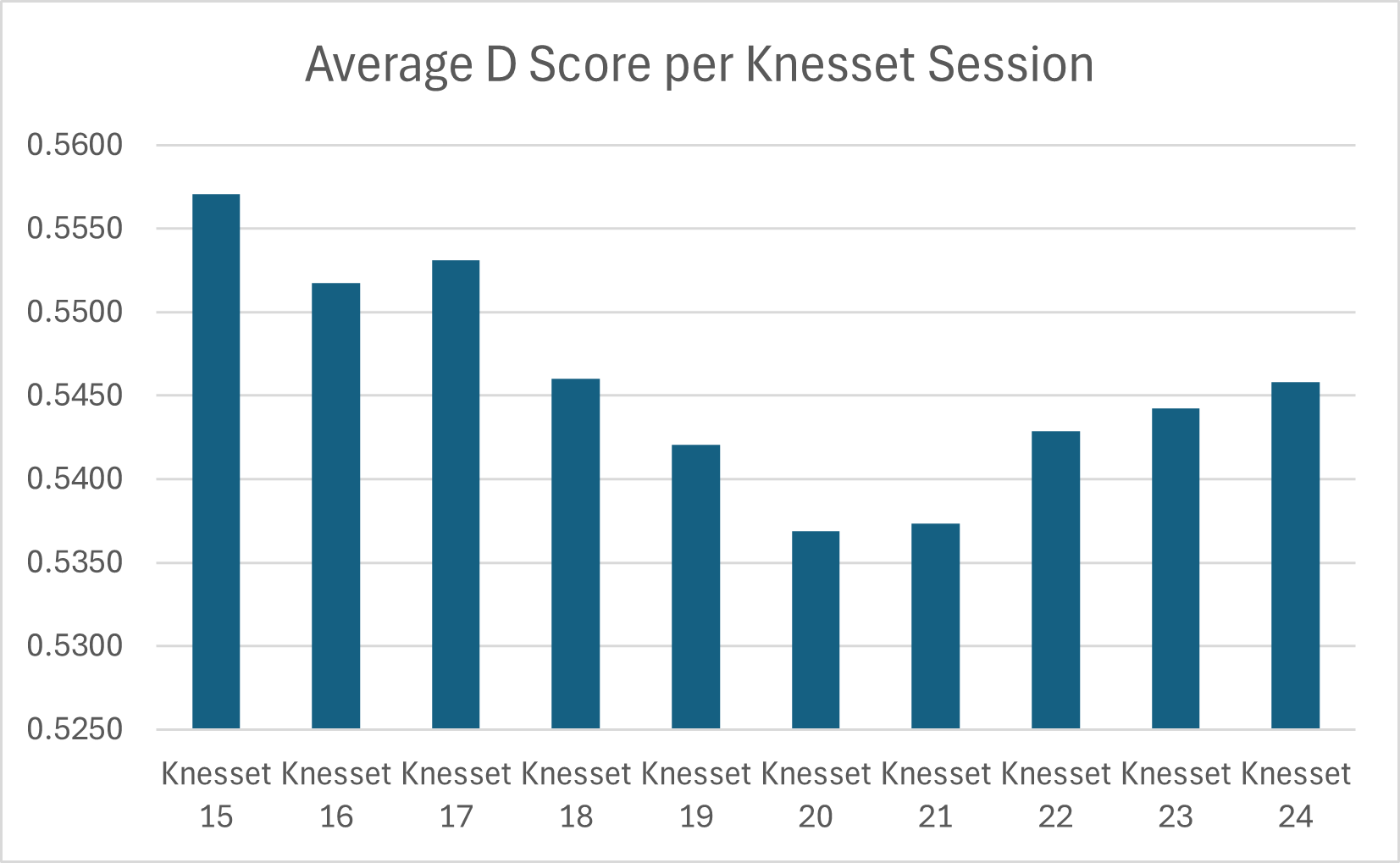}
\caption{Average Dominance score over all sentences in each Knesset session.}
\label{fig:avg_d_per_knesset}   
\end{figure}

In this section we present visualizations of VAD values in the Knesset data. 
The average V, A and D scores over all sentences in each Knesset session are presented in figures~\ref{fig:avg_v_per_knesset}, \ref{fig:avg_a_per_knesset} and \ref{fig:avg_d_per_knesset} respectively. 
More graphs, including ones that depict the proportion of high- and low-valued V, A, and D sentences by committee and as a function of time, are provided in \href{https://github.com/HaifaCLG/Polarization/}{our supplementary material}.

\section{Full OLS Regression Results tables}
\label{app:full_ols}
Tables~\ref{app:full_ols_v_var} and Table~\ref{app:full_ols_a_mean} present the full Ordinary Least Squares (OLS) regression results for the dependent variables \texttt{V\_var} and \texttt{A\_mean}, respectively. The tables include coefficients ($\beta$), standard errors (std err), t-values (t), p-values (P>|t|), and confidence intervals at the 95\% level ([0.025, 0.975]) for each predictor. Predictors include the committee names (\texttt{Comm}) and their interaction terms with the time point (\texttt{TP}), with the proportion of female members (\texttt{RatioF}) and with the proportion of government members (\texttt{RatioG}). 

\begin{table*}[hbt]
\centering
\resizebox{\textwidth}{!}{
\begin{tabular}{lcccccc}
\toprule
\textbf{Variable} & $\beta{\times}100$ & Std Err & t & P>|t| &   [0.025 &   0.975] \\
\midrule
\texttt{Comm}[Immigration, Absorption and Diaspora] &     2.24 &  8.41e-05 &  266.539 &  0.000 &    0.022 &    0.023 \\
\texttt{Comm}[Rights of the Child] &     2.31 &      0.000 &  183.787 &  0.000 &    0.023 &    0.023 \\
\texttt{Comm}[Constitution, Law and Justice] &     2.07 &  4.92e-05 &  421.127 &  0.000 &    0.021 &    0.021 \\
\texttt{Comm}[Economic Affairs] &     2.13 &  5.34e-05 &  399.059 &  0.000 &    0.021 &    0.021 \\
\texttt{Comm}[Education, Culture and Sports] &     2.31 &  5.03e-05 &  460.399 &  0.000 &    0.023 &    0.023 \\
\texttt{Comm}[Finance] &     2.22 &  4.78e-05 &  463.751 &  0.000 &    0.022 &    0.022 \\
\texttt{Comm}[Internal Affairs and Environment] &     2.20 &  5.03e-05 &  437.590 &  0.000 &    0.022 &    0.022 \\
\texttt{Comm}[House] &     2.04 &  7.47e-05 &  273.225 &  0.000 &    0.020 &    0.021 \\
\texttt{Comm}[Public Petitions] &     2.23 &      0.000 &  172.465 &  0.000 &    0.022 &    0.023 \\
\texttt{Comm}[State Control] &     2.17 &      0.000 &  213.961 &  0.000 &    0.021 &    0.022 \\
\texttt{Comm}[Science and Technology] &     2.16 &      0.000 &  179.588 &  0.000 &    0.021 &    0.022 \\
\texttt{Comm}[Status of Women and Gender Equality] &     2.25 &      0.000 &  131.601 &  0.000 &    0.022 &    0.023 \\
\texttt{Comm}[Labor, Welfare and Health] &     2.18 &  4.54e-05 &  480.410 &  0.000 &    0.022 &    0.022 \\
\texttt{Comm}[Immigration, Absorption and Diaspora]:\texttt{TP} &     0.02 &  7.93e-05 &    2.121 &  0.034 &  1.27e-05 &    0.000 \\
\texttt{Comm}[Rights of the Child]:\texttt{TP} &     0.04 &  9.76e-05 &    4.150 &  0.000 &    0.000 &    0.001 \\
\texttt{Comm}[Constitution, Law and Justice]:\texttt{TP} &     0.02 &  4.52e-05 &    5.416 &  0.000 &    0.000 &    0.000 \\
\texttt{Comm}[Economic Affairs]:\texttt{TP} &     0.04 &  4.55e-05 &    9.106 &  0.000 &    0.000 &    0.001 \\
\texttt{Comm}[Education, Culture and Sports]:\texttt{TP} &     0.03 &  5.04e-05 &    6.859 &  0.000 &    0.000 &    0.000 \\
\texttt{Comm}[Finance]:\texttt{TP} &     0.06 &  4.52e-05 &   13.288 &  0.000 &    0.001 &    0.001 \\
\texttt{Comm}[Internal Affairs and Environment]:\texttt{TP} &     0.04 &  5.11e-05 &    7.675 &  0.000 &    0.000 &    0.000 \\
\texttt{Comm}[House]:\texttt{TP} &     0.09 &    6.2e-05 &   14.706 &  0.000 &    0.001 &    0.001 \\
\texttt{Comm}[Public Petitions]:\texttt{TP} &     0.04 &      0.000 &    2.871 &  0.004 &    0.000 &    0.001 \\
\texttt{Comm}[State Control]:\texttt{TP} &     0.06 &  8.45e-05 &    7.586 &  0.000 &    0.000 &    0.001 \\
\texttt{Comm}[Science and Technology]:\texttt{TP} &     0.03 &      0.000 &    2.201 &  0.028 &  2.98e-05 &    0.001 \\
\texttt{Comm}[Status of Women and Gender Equality]:\texttt{TP} &     0.04 &  8.85e-05 &    4.128 &  0.000 &    0.000 &    0.001 \\
\texttt{Comm}[Labor, Welfare and Health]:\texttt{TP} &     0.07 &  5.13e-05 &   13.229 &  0.000 &    0.001 &    0.001 \\
\texttt{Comm}[Immigration, Absorption and Diaspora]:\texttt{RatioF} &  0.001 &  8.13e-05 &    0.157 &  0.875 &   -0.000 &    0.000 \\
\texttt{Comm}[Rights of the Child]:\texttt{RatioF} &    -0.02 &  7.67e-05 &   -2.772 &  0.006 &   -0.000 & -6.23e-05 \\
\texttt{Comm}[Constitution, Law and Justice]:\texttt{RatioF} &     0.03 &  5.83e-05 &    5.885 &  0.000 &    0.000 &    0.000 \\
\texttt{Comm}[Economic Affairs]:\texttt{RatioF} &     0.04 &  6.68e-05 &    6.572 &  0.000 &    0.000 &    0.001 \\
\texttt{Comm}[Education, Culture and Sports]:\texttt{RatioF} &  0.001 &  5.56e-05 &    0.127 &  0.899 &   -0.000 &    0.000 \\
\texttt{Comm}[Finance]:\texttt{RatioF} &     0.05 &  7.05e-05 &    7.463 &  0.000 &    0.000 &    0.001 \\
\texttt{Comm}[Internal Affairs and Environment]:\texttt{RatioF} &     0.02 &    5.9e-05 &    2.787 &  0.005 &  4.88e-05 &    0.000 \\
\texttt{Comm}[House]:\texttt{RatioF} &     0.03 &      0.000 &    3.160 &  0.002 &    0.000 &    0.001 \\
\texttt{Comm}[Public Petitions]:\texttt{RatioF} &  0.004 & 
  7.92e-05 &    0.518 &  0.604 &   -0.000 &    0.000 \\
\texttt{Comm}[State Control]:\texttt{RatioF} &     0.03 &  7.75e-05 &    3.669 &  0.000 &    0.000 &     0.000 \\
\texttt{Comm}[Science and Technology]:\texttt{RatioF} &    -0.02 &  9.72e-05 &   -2.489 &  0.013 &   -0.000 & -5.14e-05 \\
\texttt{Comm}[Status of Women and Gender Equality]:\texttt{RatioF} &   -0.01 &  7.95e-05 &   -1.163 &  0.245 &   -0.000 &  6.34e-05 \\
\texttt{Comm}[Labor, Welfare and Health]:\texttt{RatioF} &     0.02 &  4.88e-05 &    3.919 &  0.000 &  9.56e-05 &    0.000 \\
\texttt{Comm}[Immigration, Absorption and Diaspora]:\texttt{RatioG} &    -0.03 &  8.67e-05 &   -2.908 &  0.004 &   -0.000 & -8.23e-05 \\
\texttt{Comm}[Rights of the Child]:\texttt{RatioG} &    -0.01 &  8.47e-05 &   -1.604 &  0.109 &   -0.000 &  3.01e-05 \\
\texttt{Comm}[Constitution, Law and Justice]:\texttt{RatioG} &    -0.02 &  5.25e-05 &   -3.768 &  0.000 &   -0.000 & -9.49e-05 \\
\texttt{Comm}[Economic Affairs]:\texttt{RatioG} &    -0.02 &  4.23e-05 &   -4.026 &  0.000 &   -0.000 & -8.74e-05 \\
\texttt{Comm}[Education, Culture and Sports]:\texttt{RatioG} &    -0.03 &  5.07e-05 &   -6.835 &  0.000 &   -0.000 &   -0.000 \\
\texttt{Comm}[Finance]:\texttt{RatioG} &    -0.04 &  5.65e-05 &   -7.722 &  0.000 &   -0.001 &   -0.000 \\
\texttt{Comm}[Internal Affairs and Environment]:\texttt{RatioG} &    -0.03 &  5.39e-05 &   -5.249 &  0.000 &   -0.000 &   -0.000 \\
\texttt{Comm}[House]:\texttt{RatioG} &    -0.08 &  9.13e-05 &   -8.674 &  0.000 &   -0.001 &   -0.001 \\
\texttt{Comm}[Public Petitions]:\texttt{RatioG} &    -0.02 &      0.000 &   -1.787 &  0.074 &   -0.000 &  1.77e-05 \\
\texttt{Comm}[State Control]:\texttt{RatioG} & -2.458e-03 &  8.51e-05 &   -0.289 &  0.773 &   -0.000 &    0.000 \\
\texttt{Comm}[Science and Technology]:\texttt{RatioG} & -3.334e-03 &      0.000 &   -0.330 &  0.742 &   -0.000 &    0.000 \\
\texttt{Comm}[Status of Women and Gender Equality]:\texttt{RatioG} &     0.01 &  8.05e-05 &    1.579 &  0.114 &  -3.06e-05 &    0.000 \\
\texttt{Comm}[Labor, Welfare and Health]:\texttt{RatioG} &    -0.03 &  4.26e-05 &   -6.857 &  0.000 &   -0.000 &   -0.000 \\
\bottomrule
\end{tabular}}
\caption{Full OLS regression results: \texttt{V\_var}.}
\label{app:full_ols_v_var}
\end{table*}

\begin{table*}[hbt]
\centering
\resizebox{\textwidth}{!}{
\begin{tabular}{lcccccc}
\toprule
\textbf{Variable} & $\beta{\times}100$ & Std Err & t & P>|t| & [0.025 & 0.975] \\
\midrule
\texttt{Comm}[Immigration, Absorption and Diaspora] & 39.64 & 0.001 & 545.455 & 0.000 & 0.395 & 0.398 \\
\texttt{Comm}[Rights of the Child] & 39.34 & 0.001 & 362.140 & 0.000 & 0.391 & 0.396 \\
\texttt{Comm}[Constitution, Law and Justice] & 39.62 & 0.000 & 931.937 & 0.000 & 0.395 & 0.397 \\
\texttt{Comm}[Economic Affairs] & 38.11 & 0.000 & 826.041 & 0.000 & 0.380 & 0.382 \\
\texttt{Comm}[Education, Culture and Sports] & 38.99 & 0.000 & 897.900 & 0.000 & 0.389 & 0.391 \\
\texttt{Comm}[Finance] & 38.27 & 0.000 & 927.469 & 0.000 & 0.382 & 0.383 \\
\texttt{Comm}[Internal Affairs and Environment] & 39.36 & 0.000 & 906.172 & 0.000 & 0.393 & 0.394 \\
\texttt{Comm}[House] & 40.37 & 0.001 & 625.423 & 0.000 & 0.402 & 0.405 \\
\texttt{Comm}[Public Petitions] & 39.03 & 0.001 & 350.026 & 0.000 & 0.388 & 0.392 \\
\texttt{Comm}[State Control] & 39.67 & 0.001 & 453.011 & 0.000 & 0.395 & 0.398 \\
\texttt{Comm}[Science and Technology] & 38.29 & 0.001 & 368.023 & 0.000 & 0.381 & 0.385 \\
\texttt{Comm}[Status of Women and Gender Equality] & 40.32 & 0.001 & 272.855 & 0.000 & 0.400 & 0.406 \\
\texttt{Comm}[Labor, Welfare and Health] & 38.12 & 0.000 & 973.036 & 0.000 & 0.380 & 0.382 \\
\texttt{Comm}[Immigration, Absorption and Diaspora]:\texttt{TP} & 0.55 & 0.001 & 8.040 & 0.000 & 0.004 & 0.007 \\
\texttt{Comm}[Rights of the Child]:\texttt{TP} & 0.35 & 0.001 & 4.113 & 0.000 & 0.002 & 0.005 \\
\texttt{Comm}[Constitution, Law and Justice]:\texttt{TP} & -0.33 & 0.000 & -8.542 & 0.000 & -0.004 & -0.003 \\
\texttt{Comm}[Economic Affairs]:\texttt{TP} & 0.30 & 0.000 & 7.660 & 0.000 & 0.002 & 0.004 \\
\texttt{Comm}[Education, Culture and Sports]:\texttt{TP} & 0.49 & 0.000 & 11.173 & 0.000 & 0.004 & 0.006 \\
\texttt{Comm}[Finance]:\texttt{TP} & 0.21 & 0.000 & 5.497 & 0.000 & 0.001 & 0.003 \\
\texttt{Comm}[Internal Affairs and Environment]:\texttt{TP} & 0.44 & 0.000 & 10.011 & 0.000 & 0.004 & 0.005 \\
\texttt{Comm}[House]:\texttt{TP} & 0.29 & 0.001 & 5.463 & 0.000 & 0.002 & 0.004 \\
\texttt{Comm}[Public Petitions]:\texttt{TP} & 0.32 & 0.001 & 2.912 & 0.004 & 0.001 & 0.005 \\
\texttt{Comm}[State Control]:\texttt{TP} & 0.15 & 0.001 & 2.021 & 0.043 & 0.000 & 0.003 \\
\texttt{Comm}[Science and Technology]:\texttt{TP} & 0.46 & 0.001 & 4.319 & 0.000 & 0.003 & 0.007 \\
\texttt{Comm}[Status of Women and Gender Equality]:\texttt{TP} & 0.03 & 0.001 & 0.440 & 0.660 & -0.001 & 0.002 \\
\texttt{Comm}[Labor, Welfare and Health]:\texttt{TP} & 0.29 & 0.000 & 6.482 & 0.000 & 0.002 & 0.004 \\
\texttt{Comm}[Immigration, Absorption and Diaspora]:\texttt{RatioF} & 0.04 & 0.001 & 0.568 & 0.570 & -0.001 & 0.002 \\
\texttt{Comm}[Rights of the Child]:\texttt{RatioF} & -0.24 & 0.001 & -3.601 & 0.000 & -0.004 & -0.001 \\
\texttt{Comm}[Constitution, Law and Justice]:\texttt{RatioF} & 0.37 & 0.001 & 7.410 & 0.000 & 0.003 & 0.005 \\
\texttt{Comm}[Economic Affairs]:\texttt{RatioF} & 0.39 & 0.001 & 6.676 & 0.000 & 0.003 & 0.005 \\
\texttt{Comm}[Education, Culture and Sports]:\texttt{RatioF} & 0.04 & 0.000 & 0.804 & 0.422 & -0.001 & 0.001 \\
\texttt{Comm}[Finance]:\texttt{RatioF} & 0.37 & 0.001 & 6.043 & 0.000 & 0.002 & 0.005 \\
\texttt{Comm}[Internal Affairs and Environment]:\texttt{RatioF} & -0.09 & 0.001 & -1.700 & 0.089 & -0.002 & 0.000 \\
\texttt{Comm}[House]:\texttt{RatioF} & 0.04 & 0.001 & 0.401 & 0.688 & -0.001 & 0.002 \\
\texttt{Comm}[Public Petitions]:\texttt{RatioF} & 0.17 & 0.001 & 2.432 & 0.015 & 0.000 & 0.003 \\
\texttt{Comm}[State Control]:\texttt{RatioF} & -0.05 & 0.001 & -0.674 & 0.500 & -0.002 & 0.001 \\
\texttt{Comm}[Science and Technology]:\texttt{RatioF} & -0.02 & 0.001 & -0.241 & 0.809 & -0.002 & 0.001 \\
\texttt{Comm}[Status of Women and Gender Equality]:\texttt{RatioF} & -0.001 & 0.001 & -0.012 & 0.991 & -0.001 & 0.001 \\
\texttt{Comm}[Labor, Welfare and Health]:\texttt{RatioF} & 0.14 & 0.000 & 3.238 & 0.001 & 0.001 & 0.002 \\
\texttt{Comm}[Immigration, Absorption and Diaspora]:\texttt{RatioG} & -0.24 & 0.001 & -3.225 & 0.001 & -0.004 & -0.001 \\
\texttt{Comm}[Rights of the Child]:\texttt{RatioG} & 0.20 & 0.001 & 2.753 & 0
.006 & 0.001 & 0.003 \\
\texttt{Comm}[Constitution, Law and Justice]:\texttt{RatioG} & -0.30 & 0.000 & -6.603 & 0.000 & -0.004 & -0.002 \\
\texttt{Comm}[Economic Affairs]:\texttt{RatioG} & -0.23 & 0.000 & -6.313 & 0.000 & -0.003 & -0.002 \\
\texttt{Comm}[Education, Culture and Sports]:\texttt{RatioG} & -0.22 & 0.000 & -4.960 & 0.000 & -0.003 & -0.001 \\
\texttt{Comm}[Finance]:\texttt{RatioG} & -0.43 & 0.000 & -8.850 & 0.000 & -0.005 & -0.003 \\
\texttt{Comm}[Internal Affairs and Environment]:\texttt{RatioG} & -0.37 & 0.000 & -7.944 & 0.000 & -0.005 & -0.003 \\
\texttt{Comm}[House]:\texttt{RatioG} & -0.57 & 0.001 & -7.272 & 0.000 & -0.007 & -0.004 \\
\texttt{Comm}[Public Petitions]:\texttt{RatioG} & -0.17 & 0.001 & -1.898 & 0.058 & -0.003 & 0.000 \\
\texttt{Comm}[State Control]:\texttt{RatioG} & 0.41 & 0.001 & 5.549 & 0.000 & 0.003 & 0.006 \\
\texttt{Comm}[Science and Technology]:\texttt{RatioG} & 0.06 & 0.001 & 0.710 & 0.478 & -0.001 & 0.002 \\
\texttt{Comm}[Status of Women and Gender Equality]:\texttt{RatioG} & 0.09 & 0.001 & 1.319 & 0.187 & -0.000 & 0.002 \\
\texttt{Comm}[Labor, Welfare and Health]:\texttt{RatioG} & -0.03 & 0.000 & -0.884 & 0.376 & -0.001 & 0.000 \\
\bottomrule
\end{tabular}}
\caption{Full OLS regression results: \texttt{A\_mean}.}
\label{app:full_ols_a_mean}
\end{table*}

\end{document}